\documentclass[11pt, a4paper,logo,  copyright, nonumbering]{paddleocr}
\usepackage[numbers]{natbib}
\usepackage{dblfloatfix}
\usepackage{ulem}
\usepackage{float}
\usepackage{tocloft}
\usepackage{caption}
\usepackage{dramatist}
\usepackage{xspace}
\usepackage{array}
\usepackage{pifont} %
\usepackage{multirow}
\usepackage{geometry}
\usepackage{makecell}
\usepackage{tcolorbox}
\usepackage{xltabular}
\usepackage{titlesec} 
\usepackage{longtable}
\usepackage[hang, flushmargin]{footmisc}
\usepackage{booktabs} 
\usepackage{xcolor}   
\usepackage[table]{xcolor}
\usepackage{etoolbox}
\usepackage{url}
\usepackage{changepage}
\definecolor{lightpink}{RGB}{255, 182, 193}
\definecolor{iccvblue}{rgb}{0.21,0.49,0.74}

\hypersetup{
    colorlinks=true,
    linkcolor=blue,
    citecolor=blue,
    filecolor=blue,
    urlcolor=blue,
    pdfpagemode=UseNone,
    pdfstartview=FitH
}
\usepackage{threeparttable} 
\usepackage{soul}
\usepackage{amsfonts}
\usepackage{amsmath}
\usepackage{amssymb}
\usepackage{lineno}
\usepackage{multirow}
\usepackage{adjustbox}

\usepackage{CJKutf8}
\usepackage{subcaption}
\usepackage{setspace}

\usepackage{dsfont}
\usepackage{array} %
\usepackage{tabularx} %
\usepackage{xcolor} %
\usepackage{tabularx}
\usepackage{booktabs}

\usepackage{lipsum}  %
\usepackage{multicol} %
\usepackage{makecell} 
\usepackage{listings}
\usepackage{xcolor}
\usepackage{eccvabbrv}

\usepackage{tablefootnote}
\usepackage{hyperref}

\lstdefinelanguage{json}{
    basicstyle=\ttfamily\footnotesize,
    numbers=left,
    numberstyle=\tiny\color{gray},
    stepnumber=1,
    numbersep=8pt,
    showstringspaces=false,
    breaklines=true,
    frame=lines,
    backgroundcolor=\color{gray!10},
    morestring=[b]",
    literate=
     *{0}{{{\color{black}0}}}{1}
      {1}{{{\color{black}1}}}{1}
      {2}{{{\color{black}2}}}{1}
      {3}{{{\color{black}3}}}{1}
      {4}{{{\color{black}4}}}{1}
      {5}{{{\color{black}5}}}{1}
      {6}{{{\color{black}6}}}{1}
      {7}{{{\color{black}7}}}{1}
      {8}{{{\color{black}8}}}{1}
      {9}{{{\color{black}9}}}{1}
}

\makeatletter
\def\@BTrule[#1]{%
  \ifx\longtable\undefined
    \let\@BTswitch\@BTnormal
  \else\ifx\hline\LT@hline
    \nobreak
    \let\@BTswitch\@BLTrule
  \else
     \let\@BTswitch\@BTnormal
  \fi\fi
  \global\@thisrulewidth=#1\relax
  \ifnum\@thisruleclass=\tw@\vskip\@aboverulesep\else
  \ifnum\@lastruleclass=\z@\vskip\@aboverulesep\else
  \ifnum\@lastruleclass=\@ne\vskip\doublerulesep\fi\fi\fi
  \@BTswitch}
\makeatother

\addto\extrasenglish{
}

 {\begin{list}{}%
         {\setlength{\leftmargin}{#1}}%
         \item[]%
 }
 {\end{list}}

\reportnumber{001} %

\renewcommand{\today}{}

\title{\centering Real5-OmniDocBench: A Full-Scale Physical Reconstruction Benchmark for Robust Document Parsing in the Wild}

\author[1,*]{
\small
Cheng Cui\textsuperscript{1}, Changda Zhou\textsuperscript{1}, Ziyue Gao\textsuperscript{1,2},  
\vspace{-0.3cm}
\\
\small
Xueqing Wang\textsuperscript{1},Tingquan Gao\textsuperscript{1}, Jing Tang\textsuperscript{2}, Yi Liu\textsuperscript{1} 
\vspace{0.2cm}
\\
\small
\textsuperscript{1}\textbf{PaddlePaddle Team, Baidu Inc.} 
\\
\small
\textsuperscript{2}\textbf{The Hong Kong University of Science and Technology (Guangzhou)}
\vspace{0.2cm}
\\
\small
\texttt{paddleocr@baidu.com}
\vspace{-0.5cm}
}

\renewcommand{\phi}{\varphi}

\renewcommand{\epsilon}{\varepsilon}
\renewcommand{\imath}{\mathrm{i}}

\newlength{\restsubwidth}
\newlength{\restsubheight}
\newlength{\restsubmoreheight}
\newcommand{\rest}[2]{%
        \settowidth{\restsubwidth}{\ensuremath{#2}}
        \settoheight{\restsubheight}{\ensuremath{{}_{#2}}}
        \ensuremath{{#1\hskip 0.5pt}_{\vrule\kern2pt\parbox[b][%
        4pt][b]{\the\restsubwidth}{%
                        \ensuremath{{}_{#2}}}}}
        }

\begin{abstract}
\vspace{-0.3cm} 
While Vision-Language Models (VLMs) achieve near-perfect scores on digital document benchmarks like OmniDocBench, their performance in the unpredictable physical world remains largely unknown due to the lack of controlled yet realistic evaluations. We introduce Real5-OmniDocBench, the first benchmark that performs a full-scale, one-to-one physical reconstruction of the entire OmniDocBench v1.5 (1,355 images) across five critical real-world scenarios: Scanning, Warping, Screen-Photography, Illumination, and Skew. Unlike prior benchmark that either lack digital correspondence or employ partial sampling, our complete ground-truth mapping enables, for the first time, rigorous factor-wise attribution of performance degradation-allowing us to pinpoint whether failures stem from geometric distortions, optical artifacts, or model limitations. Our benchmark establishes a challenging new standard for the community, demonstrating that the ``reality gap'' in document parsing is far from closed, and provides a diagnostic tool to guide the development of truly resilient document intelligence.

\end{abstract}

\begin{document}

\maketitle

\let\thefootnote\relax\footnote{%
  \includegraphics[height=1.0em]{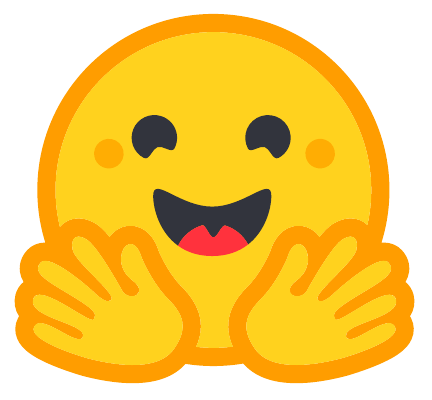}~%
  The dataset is publicly available at: \url{https://huggingface.co/datasets/PaddlePaddle/Real5-OmniDocBench}%
}
\vspace{0cm}

\section{Introduction}
\label{sec:intro}

Document parsing, defined as the task of transforming unstructured document images into structured formats such as Markdown and LaTeX, has become a critical benchmark for evaluating the fine-grained visual reasoning capabilities of Vision-Language Models (VLMs). Recent progress has been largely driven by high-quality benchmarks like OmniDocBench \cite{omnidocbench2024}, which provide precise annotations across diverse document types and enable standardized evaluation. However, these benchmarks predominantly rely on "born-digital" documents captured under ideal, distortion-free conditions. The uncomfortable truth is that a model scoring extremely high on OmniDocBench today might fail catastrophically when faced with real-world documents, such as a page curved by a book spine, a receipt photographed under a desk lamp, or a screenshot corrupted by moiré patterns.

In real-world deployment, document images inevitably suffer from complex physical perturbations: non-rigid warping from binding, perspective distortions from handheld capture, non-uniform illumination from ambient light, and optical artifacts from secondary screen photography. Despite anecdotal evidence of such vulnerabilities, the community lacks a systematic understanding of how and why models fail under these conditions. Prior evaluation efforts fall into two categories, both insufficient: uncontrolled in-the-wild datasets (\eg, WildDoc \cite{wilddoc2025}) capture realistic degradations but lack digital ground-truth correspondence, making it impossible to precisely measure the impact of real-world environmental interference on model reasoning;  partial physical simulations (\eg, DocPTBench \cite{docptbench2025}) offer controllability but employ coarse-grained sampling that fails to capture the full spectrum of real-world distortions or enable rigorous factor-wise diagnosis. What is missing is a benchmark that combines the realism of physical capture with the precision of controlled one-to-one digital mapping—a tool that can tell us not just that a model failed but also exactly which distortion caused the failure.

We introduce Real5-OmniDocBench to fill this critical gap. Our core innovation is simple yet powerful: we perform a full-scale, one-to-one physical reconstruction of the entire OmniDocBench v1.5 test set, consisting of 1,355 pages, across five meticulously designed physical scenarios, including Scanning, Warping, Screen-Photography, Illumination, and Skew. For each digital source image, we produce five physical variants using professional-grade printing and heterogeneous mobile capture devices while inheriting the complete ground-truth annotations from the original. This design transforms physical distortions from uncontrolled confounders into controlled, independent variables. Consequently, researchers can perform causal analysis of the domain gap between ideal conditions and real-world physical scenarios for the first time, precisely evaluating performance variations across different scenarios under fully comparable settings. This achieves a level of comparability that was previously unattainable.

Our contributions are threefold:

\begin{itemize}
    \item \textbf{First Full-Scale Physical Benchmark for Causal Robustness Analysis.} We present Real5-OmniDocBench, the first benchmark offering a complete 1,355-image physical reconstruction with strict one-to-one digital correspondence. Unlike partial or uncontrolled datasets, our design enables rigorous attribution of performance degradation to specific physical factors, providing the community with a diagnostic tool rather than just another leaderboard.
    
    \item \textbf{Comprehensive Multi-Scenario Physical Modeling.} We systematically construct five orthogonal physical Scanning, Warping, Screen-Photography, Illumination, and Skew built from multiple diverse sub-conditions (\eg, Warping: folding, cylinder, crumpling, corner, book). This scenario-level design enables precise diagnosis of model robustness—for instance, revealing that a model resilient to scanning artifacts degrades severely under handheld warping.
    
    \item \textbf{Large-Scale Empirical Study Revealing Counter-Intuitive Insights.} Through comprehensive benchmarking of 17 state-of-the-art models, we uncover a finding that challenges the prevailing scaling paradigm: compact, domain-specialized VLMs (\eg, PaddleOCR-VL-1.6 at 0.9B parameters) consistently outperform much larger generalist models under physical stress. This suggests that robustness to real-world distortions is not a simple function of parameter count, but requires domain-specific inductive biases—a crucial insight for future model design.
\end{itemize}

By publicly releasing Real5-OmniDocBench, we provide the community with a much-needed stress test for document intelligence. Our results establish a demanding baseline that reveals the substantial gap between digital perfection and real-world reliability. We hope this benchmark will catalyze research toward models that truly understand documents in both pristine archives and the messy, unpredictable physical world.

\section{Related Work}
\label{sec:related}

\paragraph{Evolution of Document Parsing Benchmarks.}

Document understanding benchmarks have evolved from early single-task OCR evaluations, such as the ICDAR series \cite{icdar2019} focusing on text detection and recognition, to complex end-to-end structured parsing. In this progression, OmniDocBench \cite{omnidocbench2024} has established itself as a core benchmark for evaluating the document parsing capabilities of VLMs (\eg, Qwen3-VL \cite{bai2025qwen3vltechnicalreport}, Gemini 3 Pro \cite{gemini30}), owing to its definition of nine document types and a three-tier evaluation framework (full-page, module, and attribute levels). Although OmniDocBench provides high-precision annotations, it predominantly relies on ``born-digital'' images, overlooking the quality degradation inherent in real-world physical deployment.

\paragraph{Current Landscape of VLM Evaluation Sets.}
To further explore the parsing limits of VLMs, several specialized benchmarks have recently emerged. OlmOCR-Bench \cite{olmocrbench2025} introduces unit-test-driven verification for the structural equivalence of formulas and tables, yet lacks diversity in physical scenarios. OceanOCR-Bench \cite{chen2025oceanocrgeneralocrapplication} focuses on text-image association and cross-modal alignment within complex multimodal layouts. For ``in-the-wild'' scenarios, WildDoc \cite{wilddoc2025} collects a vast array of documents from natural environments, challenging models with extreme lighting and complex backgrounds. However, datasets like WildDoc lack a one-to-one correspondence with pristine digital originals, making it impossible to precisely contrast model performance between ideal conditions and real-world environments. Furthermore, while DocPTBench \cite{docptbench2025} attempts physical photography, its utility as a standardized diagnostic tool is hindered by partial sampling and a lack of fine-grained decomposition of interference factors.

\paragraph{Uniqueness of Real5-OmniDocBench.}
Real5-OmniDocBench fills the aforementioned gaps. Unlike works focused on preprocessing and restoration (\eg, DocRes \cite{docres2024}), we emphasize end-to-end parsing performance. Compared to VLM evaluations such as OCRBench-v2 \cite{fu2025ocrbenchv2improvedbenchmark}, which remain primarily based on digital images, our work achieves ``full-set alignment'' between the digital and physical domains through a complete physical reconstruction of the OmniDocBench corpus. This design enables the systematic decomposition of five physical factors, namely Scanning, Warping, Screen-Photography, Illumination, and Skew, providing a rigorous diagnostic evaluation standard for the reliability of VLMs in real-world deployment.

\section{The Real5-OmniDocBench Benchmark}
\label{sec:dataset}

\subsection{Design Principles and Overall Architecture}
\label{subsec:design}

\begin{figure}[t]
\centering
\includegraphics[width=\linewidth]{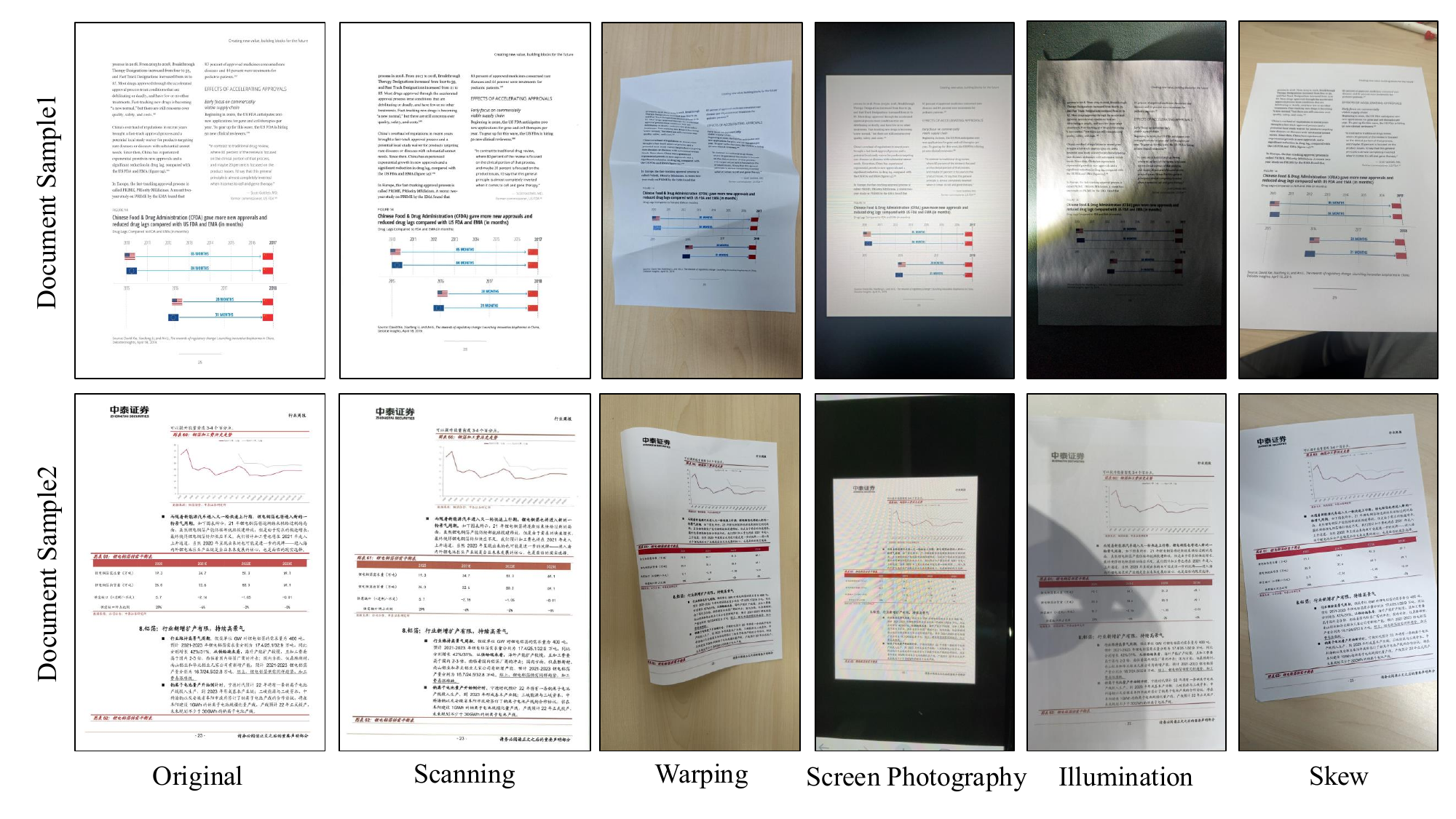}
\caption{Samples of Real5-OmniDocBench.}
\label{fig:dataset_samples}
\end{figure}

Real5-OmniDocBench is constructed based on OmniDocBench v1.5~\cite{omnidocbench2024}, which contains 1,355 pages across nine document types (academic papers, books, notes, financial reports, magazines, etc.). We follow the one-to-one mapping principle: each original sample corresponds to five real-world scenario variants, totaling 6,775 test samples. The GT inheritance strategy ensures zero-modification reuse of OmniDocBench's JSON annotations (layout, tables, formulas, text, reading order), which is the core design guaranteeing strict comparability of evaluation. Figure~\ref{fig:dataset_samples} illustrates representative samples from our dataset, showcasing the diversity of document types and real-world degradations covered.

\subsection{Physical Data Acquisition and Standardization}
\label{subsec:acquisition}

\subsection{Physical Data Acquisition and Standardization}
\label{subsec:acquisition}

To ensure the scientific integrity and reproducibility of Real5-OmniDocBench, we implement a rigorous physical data acquisition pipeline. All physical samples are reproduced using a professional-grade Canon C5840 multi-functional printer, configured at 1200\,dpi for both color printing and scanning. This high-resolution reproduction ensures that any subsequent parsing degradation is strictly attributable to environmental perturbations rather than source quality loss.

To maintain the structural fidelity of the original digital corpus, we adopt a scale-aware printing strategy: documents are categorized into A3 and A4 formats based on their native digital dimensions before printing. This approach preserves the original font-size ratios and layout densities, which are crucial for precise structural parsing evaluation. For the four handheld capture scenarios (\textit{i.e.}, Warping, Screen-Photography, Illumination, and Skew), we utilize a heterogeneous hardware matrix comprising mainstream mobile devices from Apple, Xiaomi, and OPPO. This diverse device selection is instrumental in capturing a wide spectrum of ISP (Image Signal Processing) characteristics, sensor noise profiles, and lens distortions, thereby mirroring the authentic variability of real-world user behaviors across different hardware ecosystems.

\begin{figure}[ht]
\centering
\begin{minipage}[t]{0.19\textwidth}
\centering
\includegraphics[width=\linewidth]{images/Scanning-standard.jpg} \\ \small{(a) Standard}
\end{minipage}
\hfill
\begin{minipage}[t]{0.19\textwidth}
\centering
\includegraphics[width=\linewidth]{images/Scanning-low-quality.jpg} \\ \small{(b) Low-quality}
\end{minipage}
\hfill
\begin{minipage}[t]{0.19\textwidth}
\centering
\includegraphics[width=\linewidth]{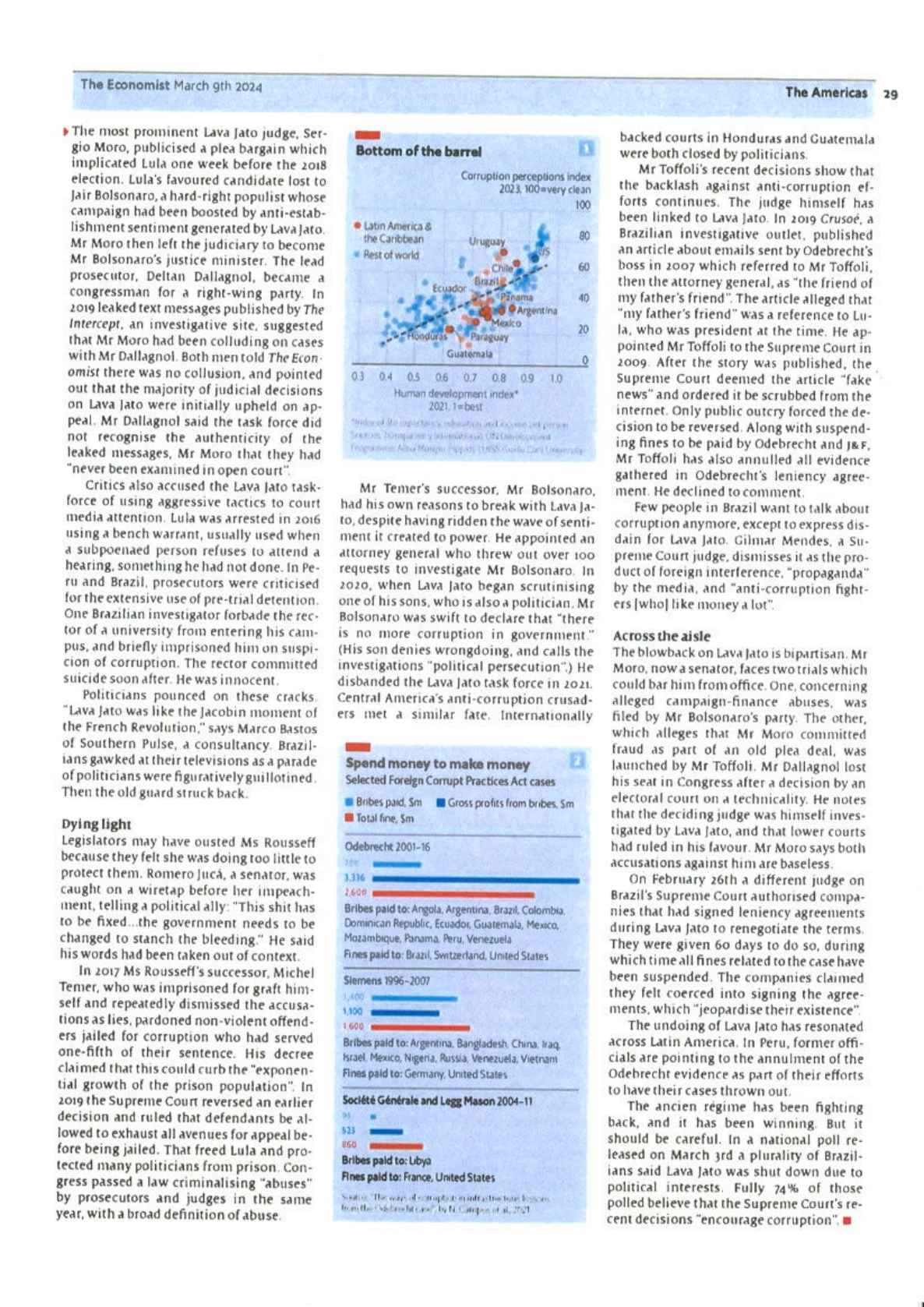} \\ \small{(c) Slanted}
\end{minipage}
\hfill
\begin{minipage}[t]{0.19\textwidth}
\centering
\includegraphics[width=\linewidth]{images/Scanning-corner-binding.jpg} \\ \small{(d) Stapled}
\end{minipage}
\hfill
\begin{minipage}[t]{0.19\textwidth}
\centering
\includegraphics[width=\linewidth]{images/Scanning-side-binding.jpg} \\ \small{(e) Bound}
\end{minipage}
\caption{
Representative configurations in the Scanning scenario.}
\label{fig:scanning_types}
\end{figure}

\paragraph{Scanning.} 
The Scanning scenario evaluates parsing performance under controlled high-resolution conditions while introducing authentic paper textures and digitization noise. As illustrated in Fig.~\ref{fig:scanning_types}, we implement five representative configurations: \textit{Standard} (a) for ideal, well-aligned capture; \textit{Low-quality} (b) simulating resolution loss from entry-level hardware via iterative print-scan cycles; \textit{Slanted} (c) capturing non-orthogonal alignment common in manual flatbed feeding; \textit{Stapled} (d) featuring local shadows and occlusions from corner clips; and \textit{Bound} (e) simulating book-style scanning with characteristic edge curvature. These settings offer a spectrum of digitization artifacts found in both archival and office environments.

\begin{figure}[ht]
\centering
\begin{minipage}[t]{0.19\textwidth}
\centering
\includegraphics[width=\linewidth]{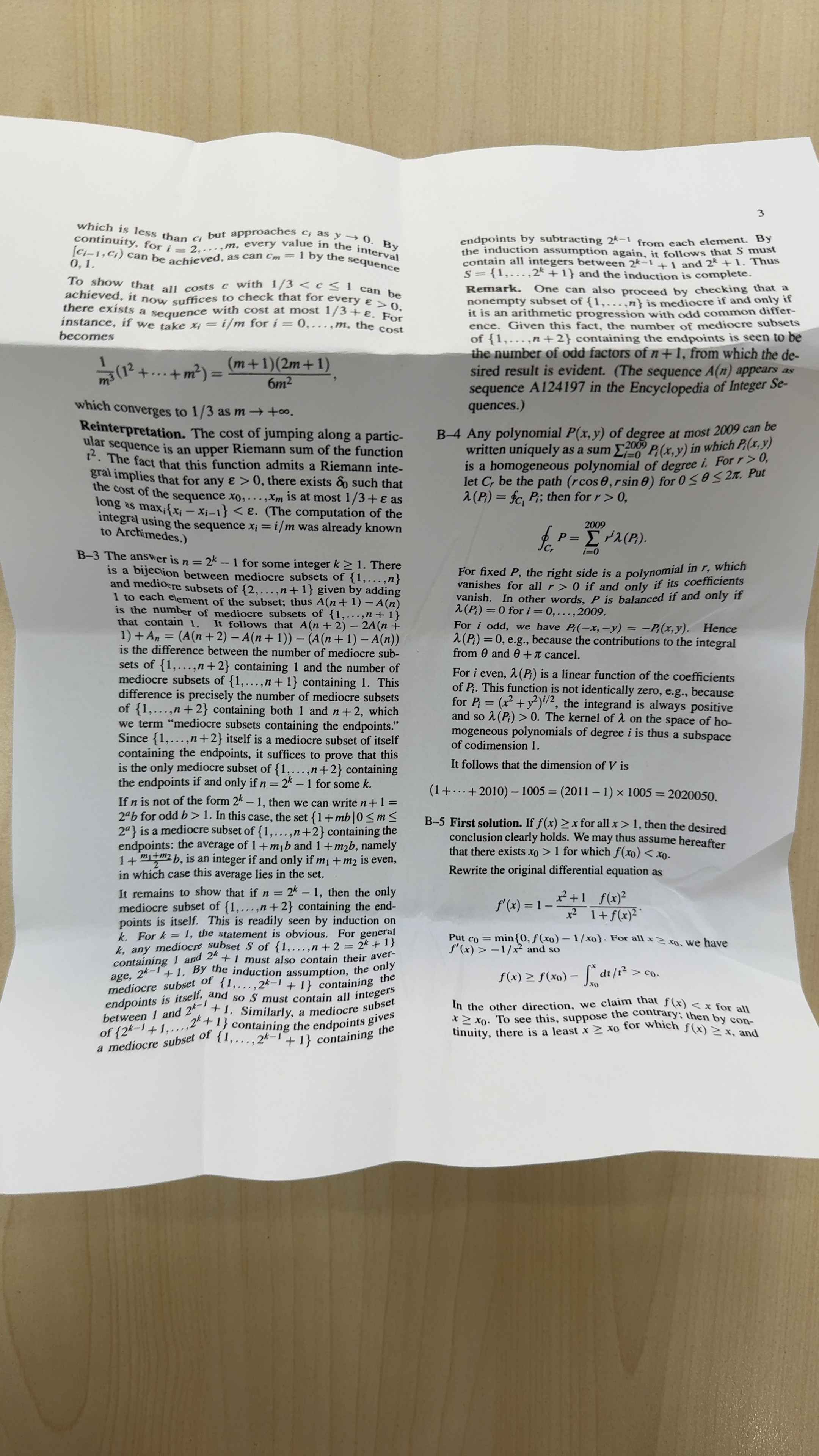} \\ \small{(a) Folding}
\end{minipage}
\hfill
\begin{minipage}[t]{0.19\textwidth}
\centering
\includegraphics[width=\linewidth]{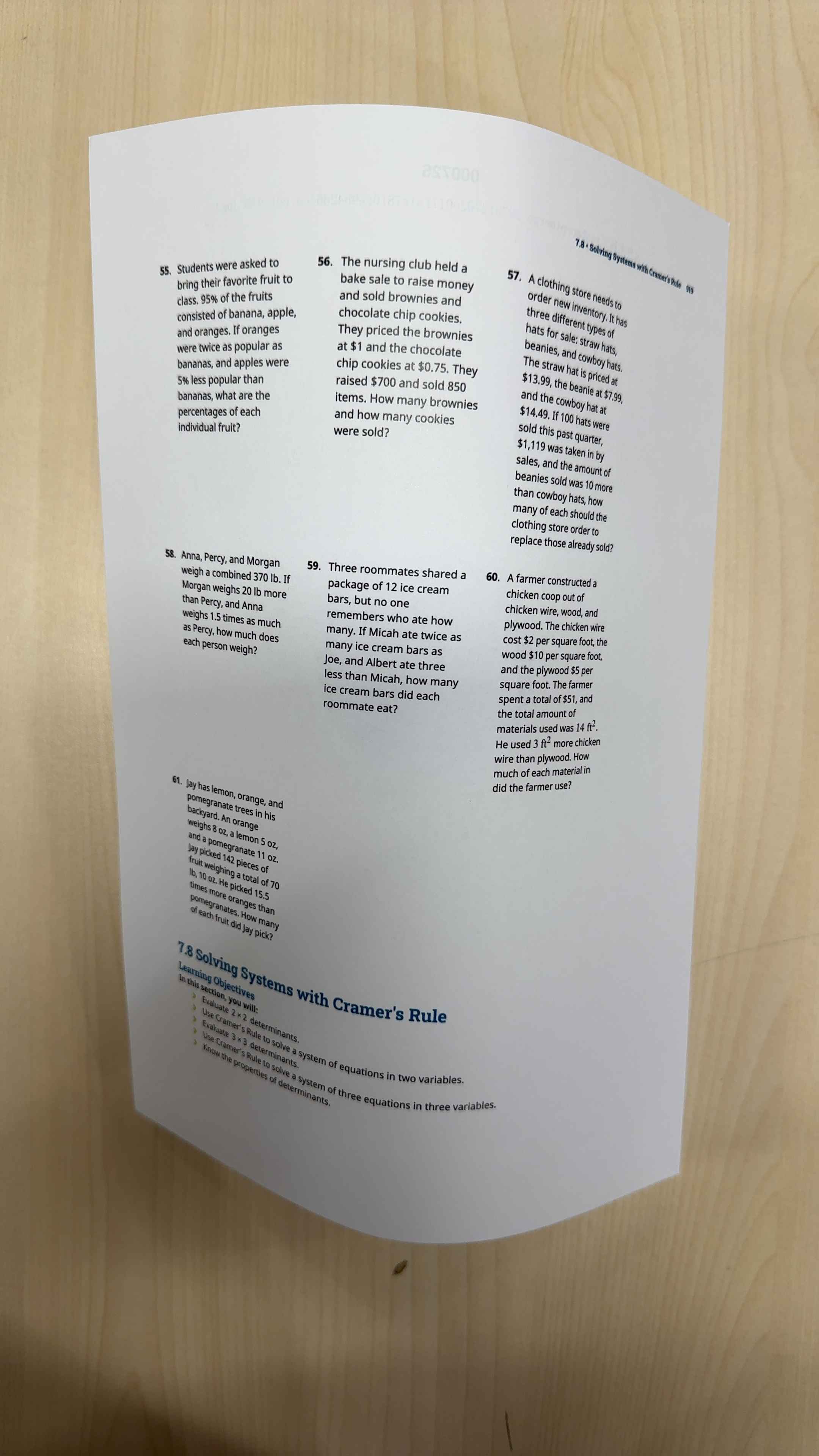} \\ \small{(b) Cylinder}
\end{minipage}
\hfill
\begin{minipage}[t]{0.19\textwidth}
\centering
\includegraphics[width=\linewidth]{images/Full-text-crumpling.jpg} \\ \small{(c) Crumpling}
\end{minipage}
\hfill
\begin{minipage}[t]{0.19\textwidth}
\centering
\includegraphics[width=\linewidth]{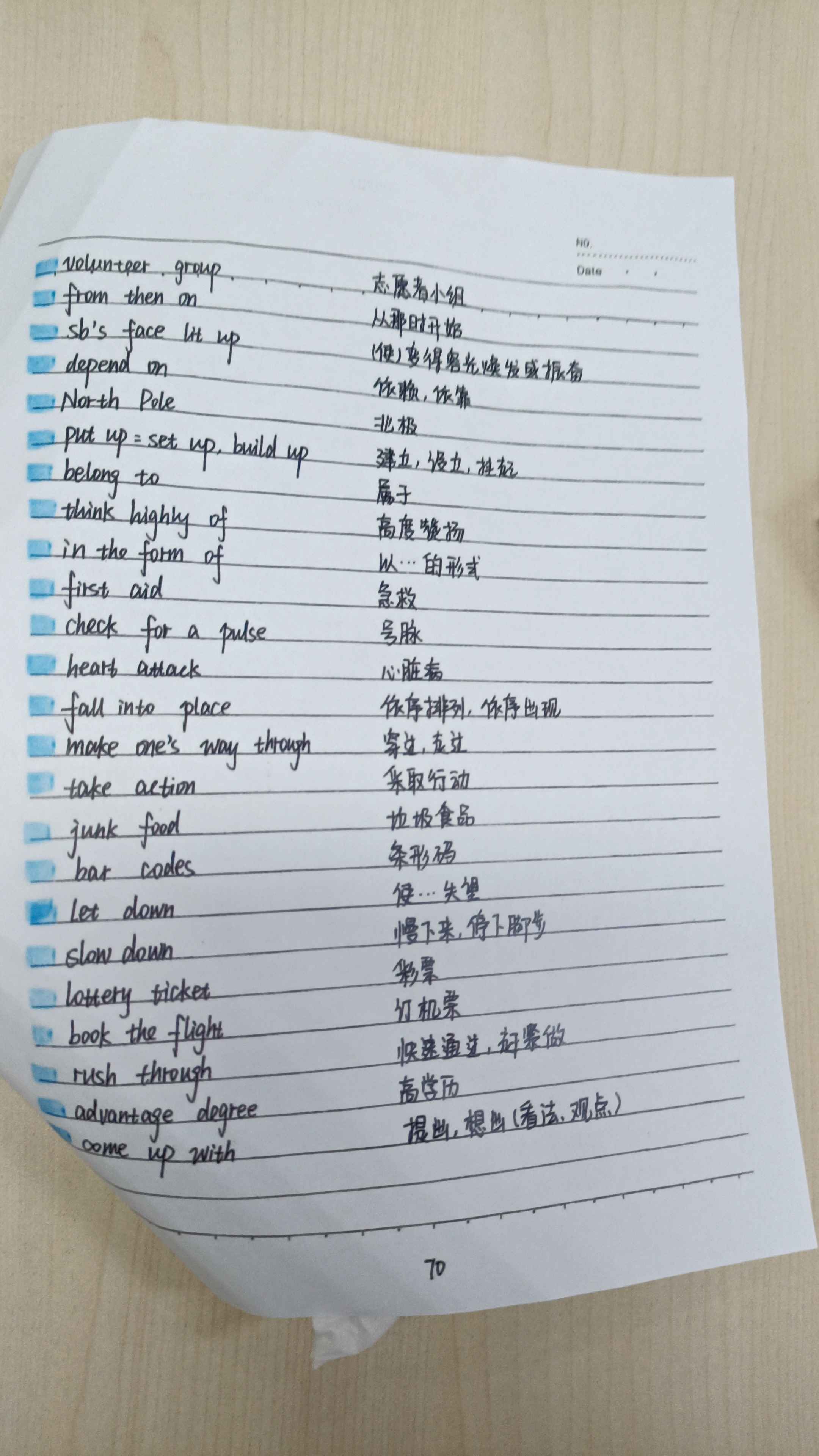} \\ \small{(d) Corner}
\end{minipage}
\hfill
\begin{minipage}[t]{0.19\textwidth}
\centering
\includegraphics[width=\linewidth]{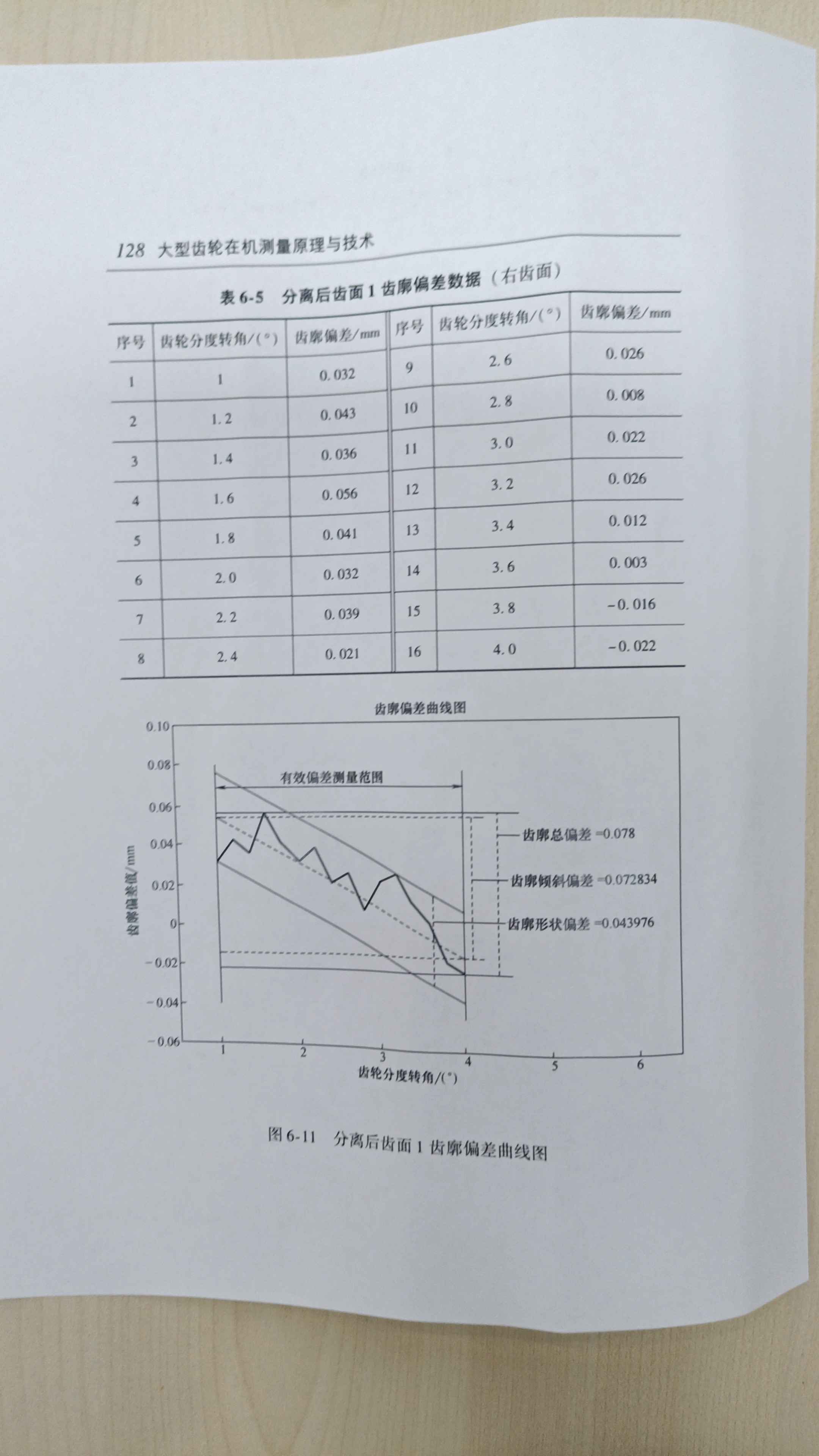} \\ \small{(e) Book}
\end{minipage}
\caption{Representative deformation types in the Warping scenario.}
\label{fig:warping_types}
\end{figure}

\paragraph{Warping.} 
The Warping scenario simulates non-rigid physical deformations encountered during document handling. We define five typical deformation patterns in Fig.~\ref{fig:warping_types}: \textit{Folding} (a) creating creases along the centerline; \textit{Cylinder} (b) producing residual curvature from rolling; \textit{Crumpling} (c) introducing dense, stochastic local folds; \textit{Corner} (d) simulating peripheral dog-ears or page curling; and \textit{Book} (e) replicating the natural arc at a binding spine. This diverse set of geometric distortions provides a rigorous test for the model's spatial robustness.

\begin{figure}[ht]
\centering
\begin{minipage}[t]{0.19\textwidth}
\centering
\includegraphics[width=\linewidth]{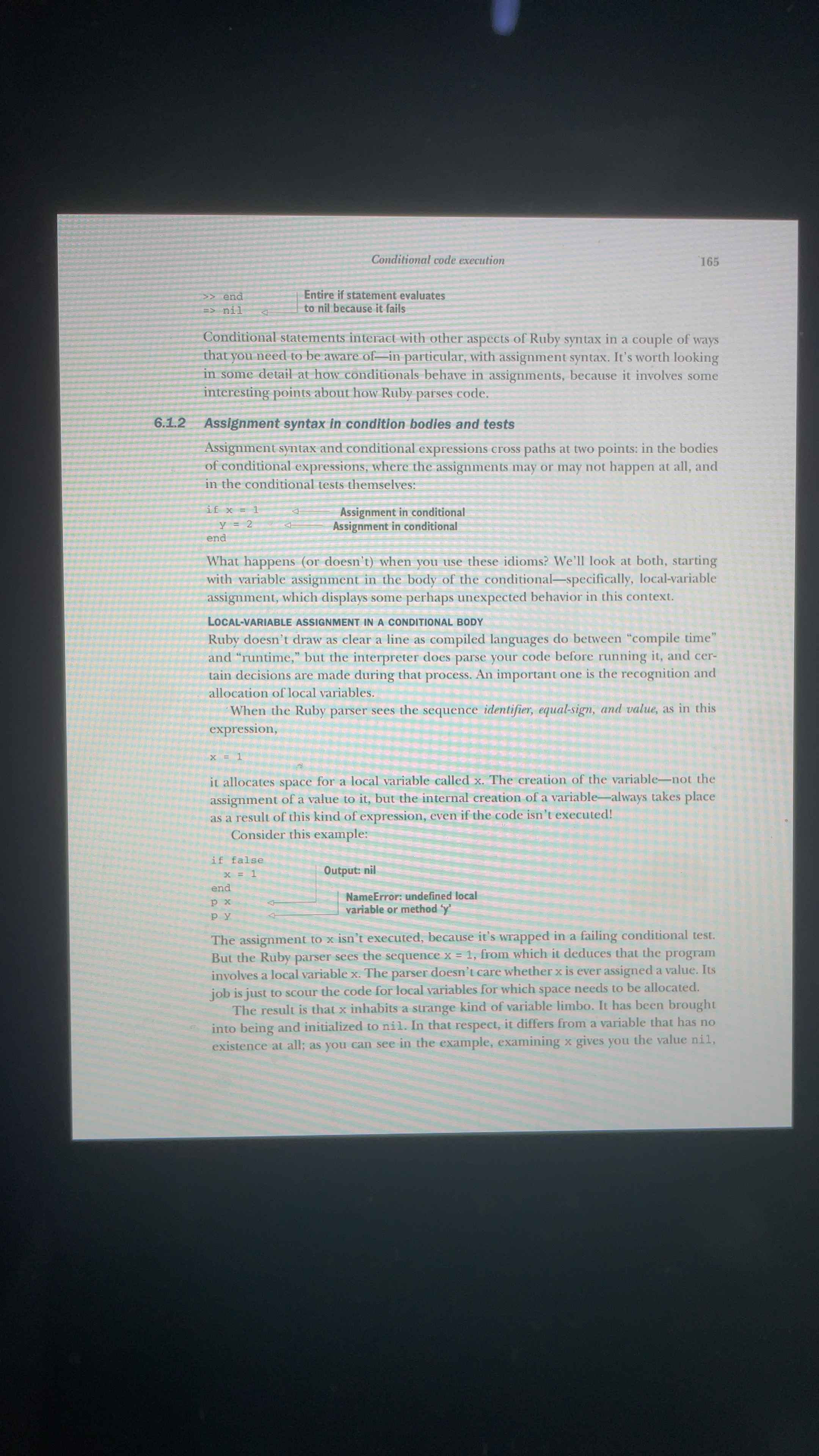} \\ \small{(a) Office}
\end{minipage}
\hfill
\begin{minipage}[t]{0.19\textwidth}
\centering
\includegraphics[width=\linewidth]{images/AOC-monitor.jpg} \\ \small{(b) Pro}
\end{minipage}
\hfill
\begin{minipage}[t]{0.19\textwidth}
\centering
\includegraphics[width=\linewidth]{images/MacBook-screen.jpg} \\ \small{(c) Laptop}
\end{minipage}
\hfill
\begin{minipage}[t]{0.19\textwidth}
\centering
\includegraphics[width=\linewidth]{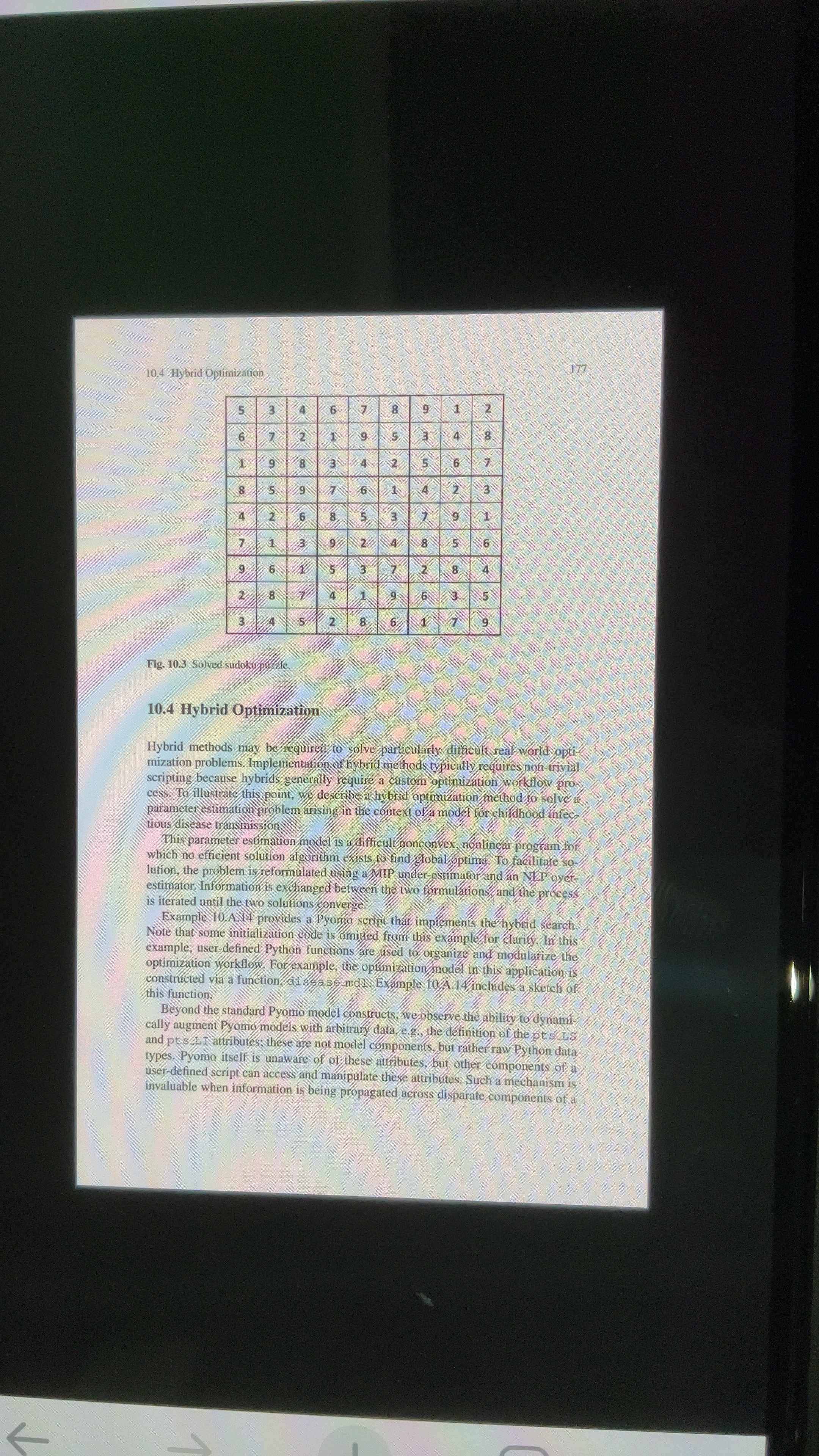} \\ \small{(d) Tablet}
\end{minipage}
\hfill
\begin{minipage}[t]{0.19\textwidth}
\centering
\includegraphics[width=\linewidth]{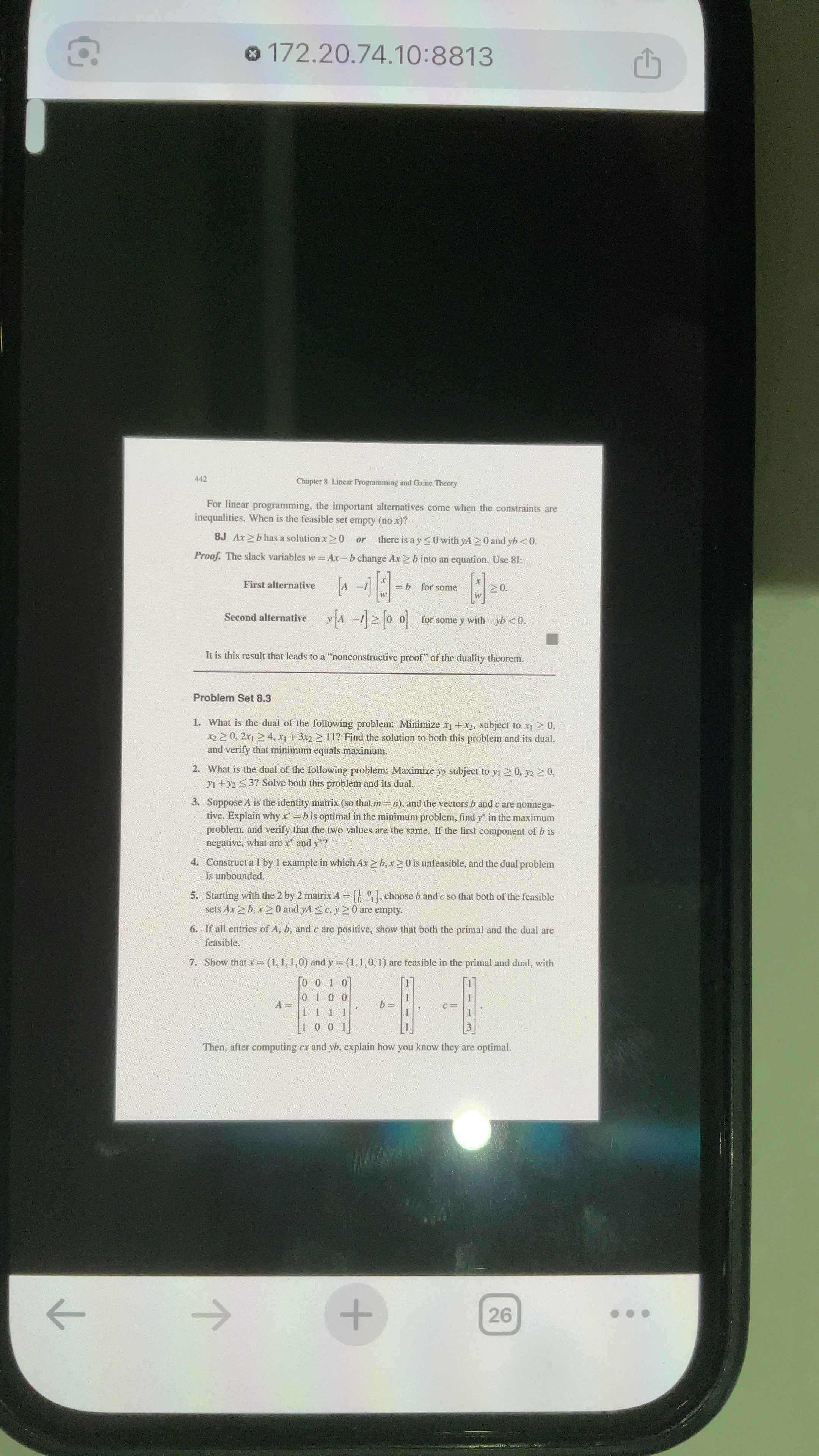} \\ \small{(e) Mobile}
\end{minipage}
\caption{Selection of display terminals for the Screen-Photography scenario.}
\label{fig:screen_types}
\end{figure}

\paragraph{Screen-Photography.} 
This scenario replicates the secondary capture of documents displayed on digital terminals. We categorize the display devices into five distinct terminals as shown in Fig.~\ref{fig:screen_types}: \textit{Office Monitor} (a) representing standard LCD displays; \textit{Professional Display} (b) featuring high pixel density; \textit{Laptop} (c) utilizing integrated high-resolution panels; \textit{Tablet} (d) simulating portable touch-screen devices; and \textit{Mobile} (e) representing small-scale, high-brightness OLED screens. This selection ensures a systematic assessment across varying pixel structures and moir\'{e} patterns.

\begin{figure}[ht]
\centering
\begin{minipage}[t]{0.19\textwidth}
\centering
\includegraphics[width=\linewidth]{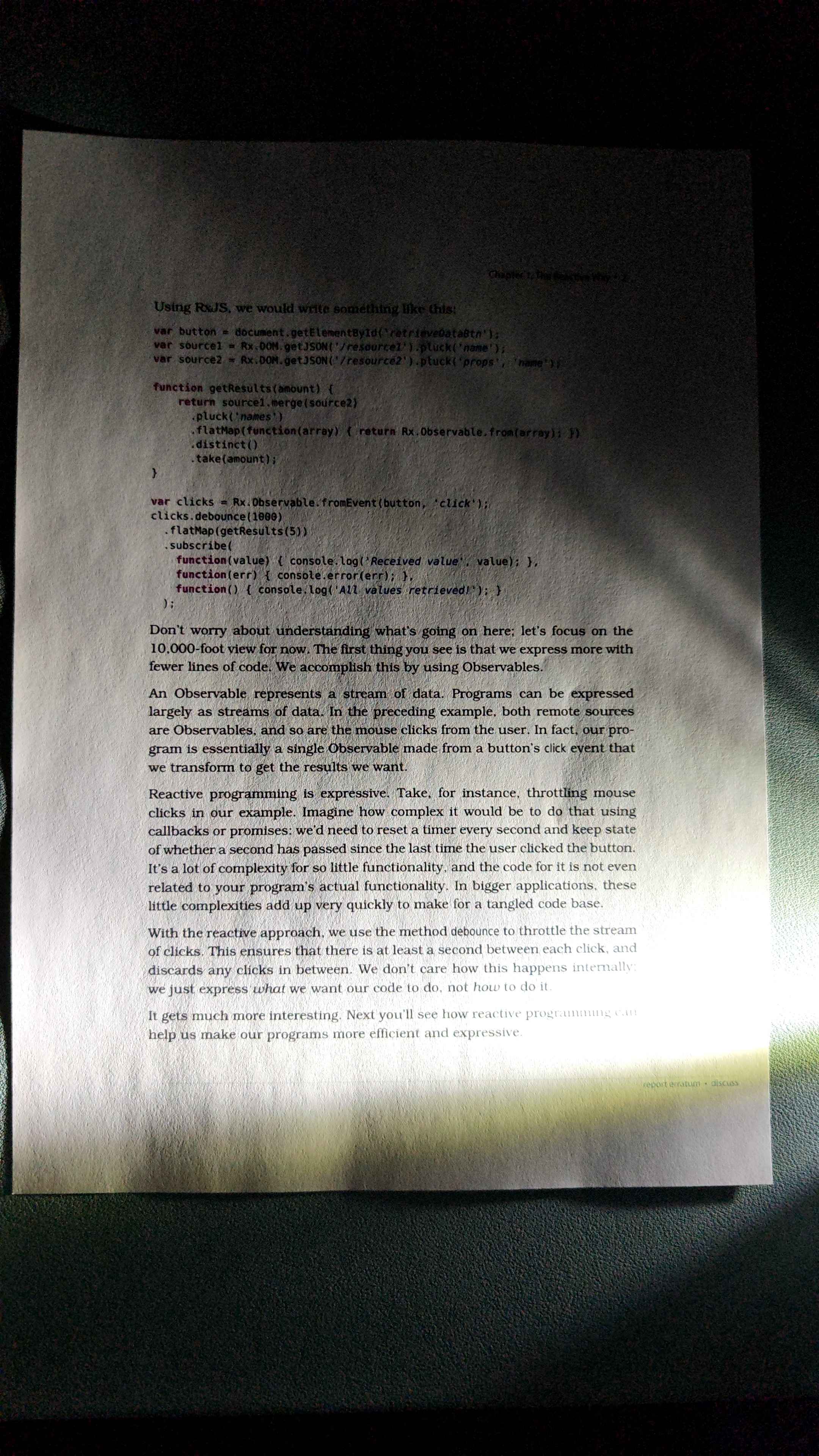} \\ \small{(a) Low-light}
\end{minipage}
\hfill
\begin{minipage}[t]{0.19\textwidth}
\centering
\includegraphics[width=\linewidth]{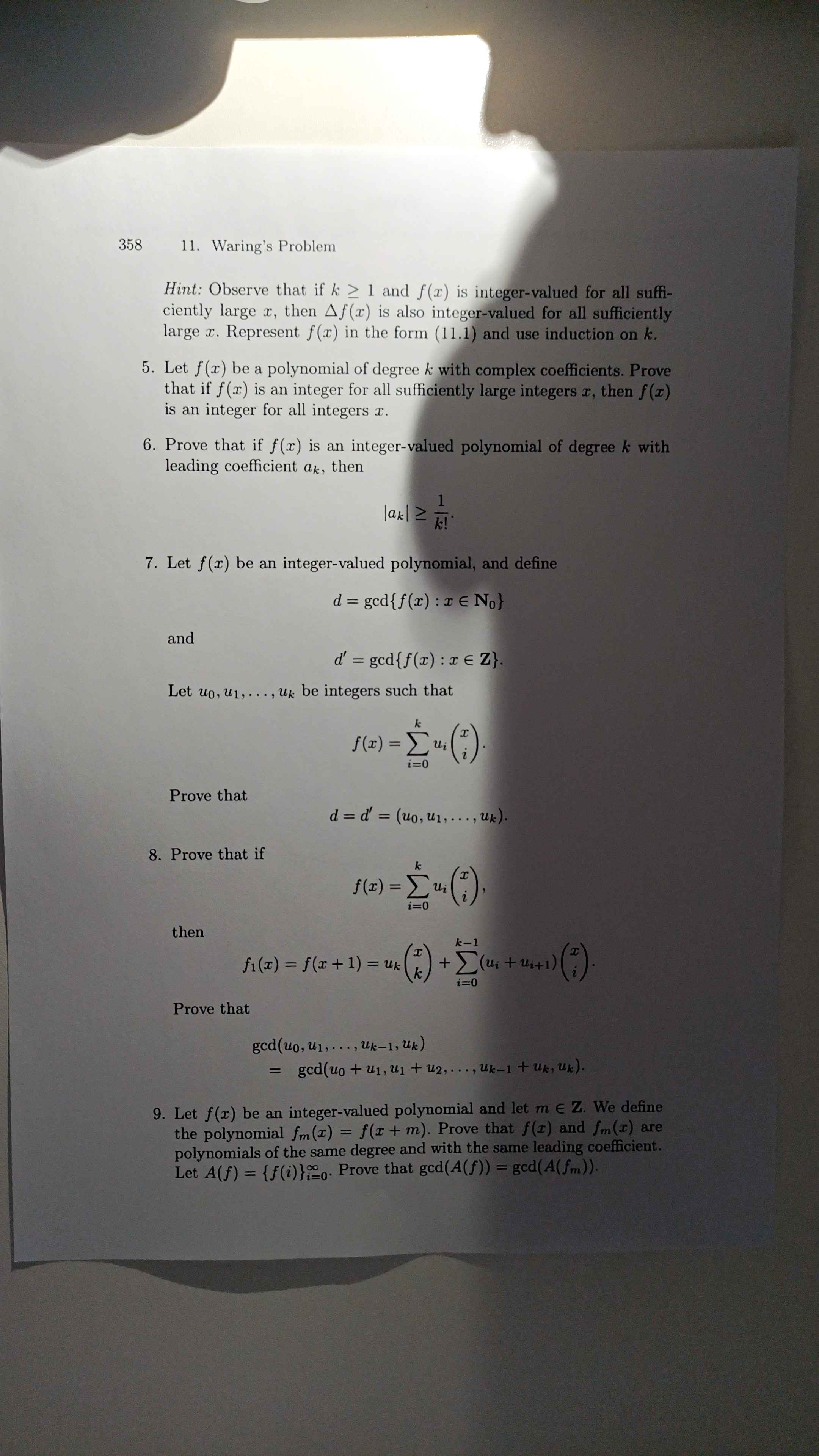} \\ \small{(b) Shadow}
\end{minipage}
\hfill
\begin{minipage}[t]{0.19\textwidth}
\centering
\includegraphics[width=\linewidth]{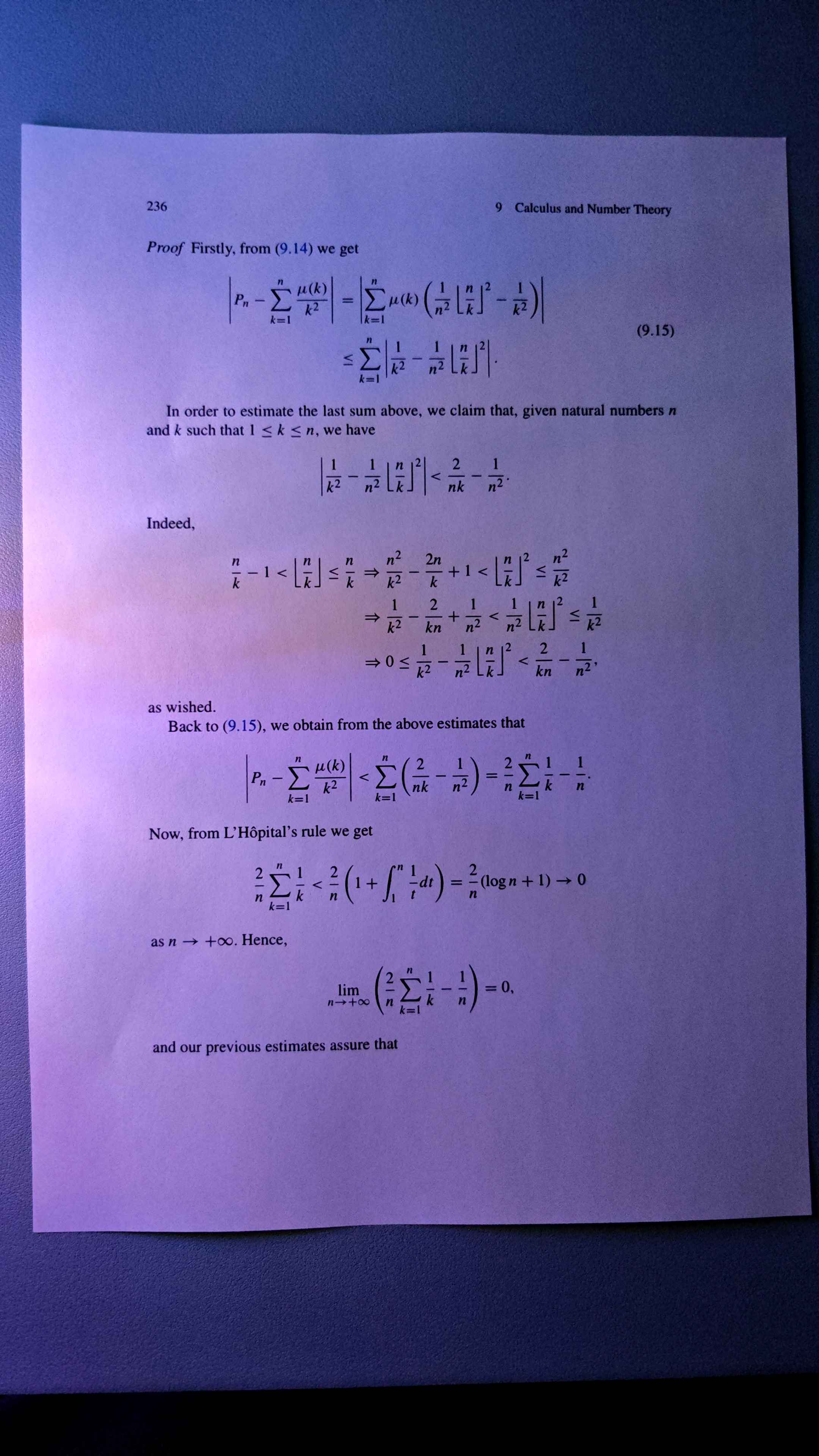} \\ \small{(c) Color-cast}
\end{minipage}
\hfill
\begin{minipage}[t]{0.19\textwidth}
\centering
\includegraphics[width=\linewidth]{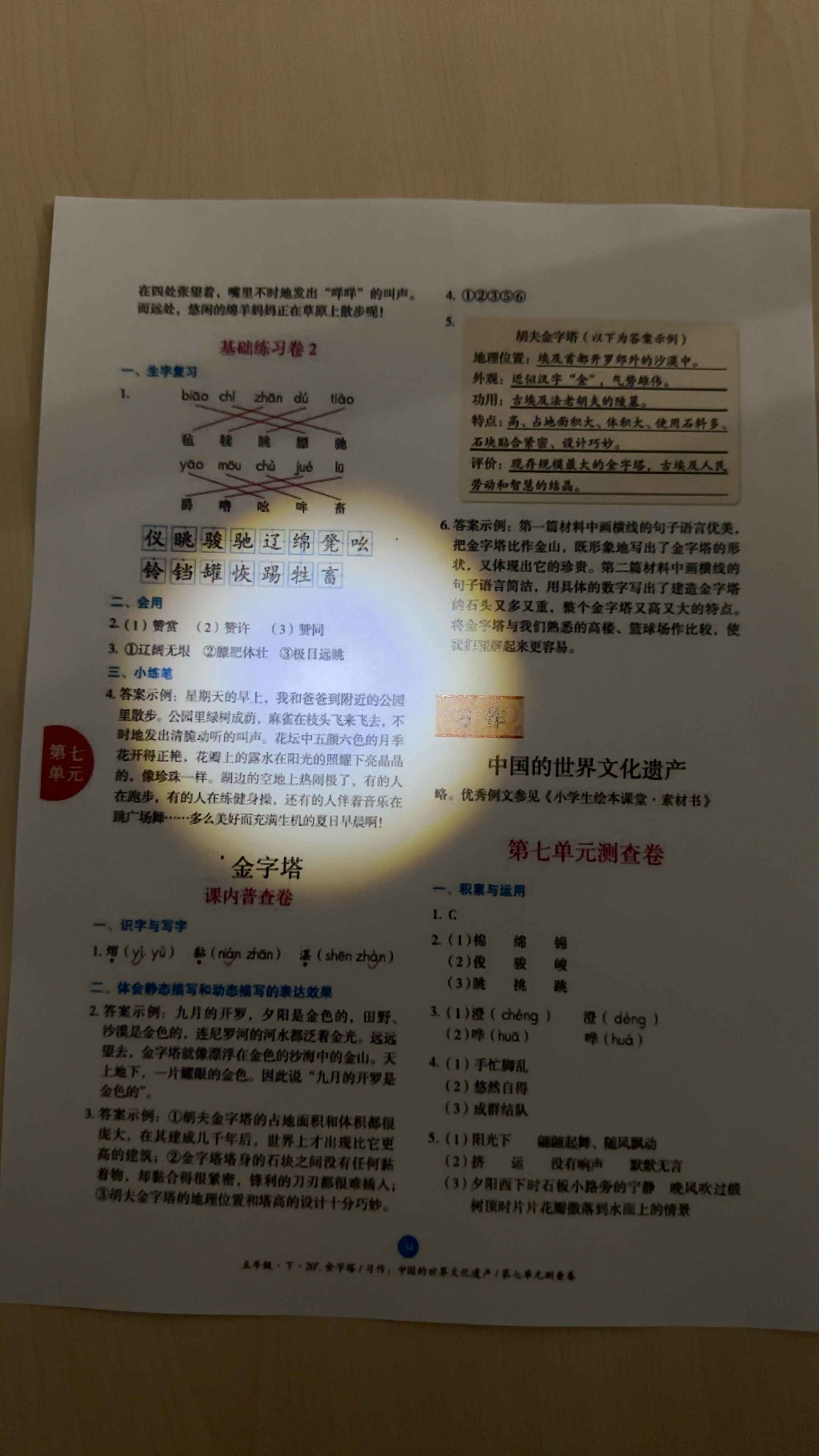} \\ \small{(d) Flashlight}
\end{minipage}
\hfill
\begin{minipage}[t]{0.19\textwidth}
\centering
\includegraphics[width=\linewidth]{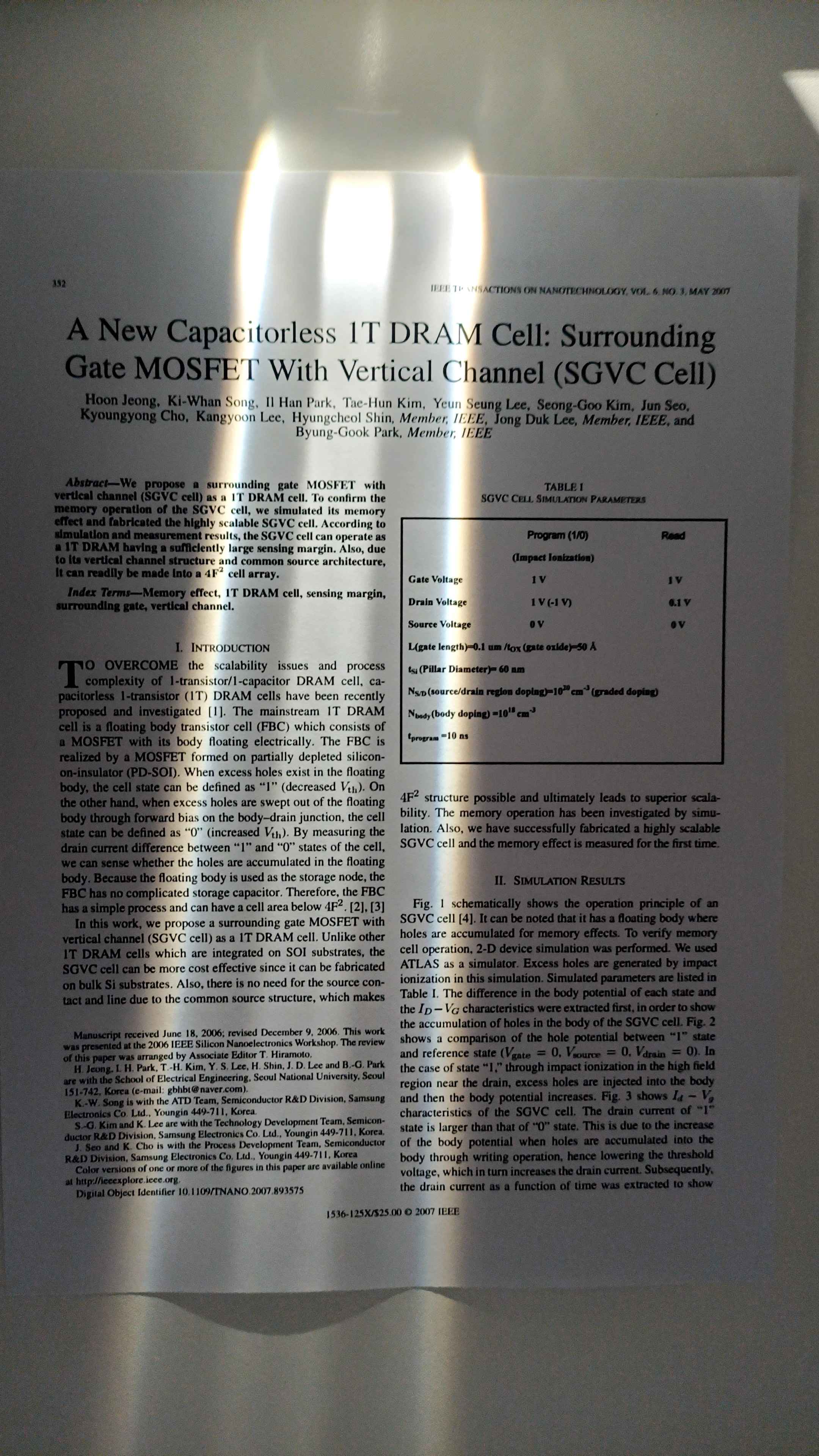} \\ \small{(e) Refraction}
\end{minipage}
\caption{Systematic lighting variations in the Illumination scenario.}
\label{fig:illumination_types}
\end{figure}

\paragraph{Illumination.} 
The Illumination scenario examines parsing resilience under non-uniform lighting. We construct five complex visual environments in Fig.~\ref{fig:illumination_types}: \textit{Low-light} (a) simulating insufficient luminance; \textit{Shadow} (b) creating partial occlusion; \textit{Color-cast} (c) introducing chromatic shifts; \textit{Flashlight} (d) generating concentrated overexposure; and \textit{Refraction} (e) simulating visual distortions through transparent media. This design probes the model's ability to maintain semantic consistency under significant contrast perturbations.

\begin{figure}[ht]
\centering
\begin{minipage}[t]{0.19\textwidth}
\centering
\includegraphics[width=\linewidth]{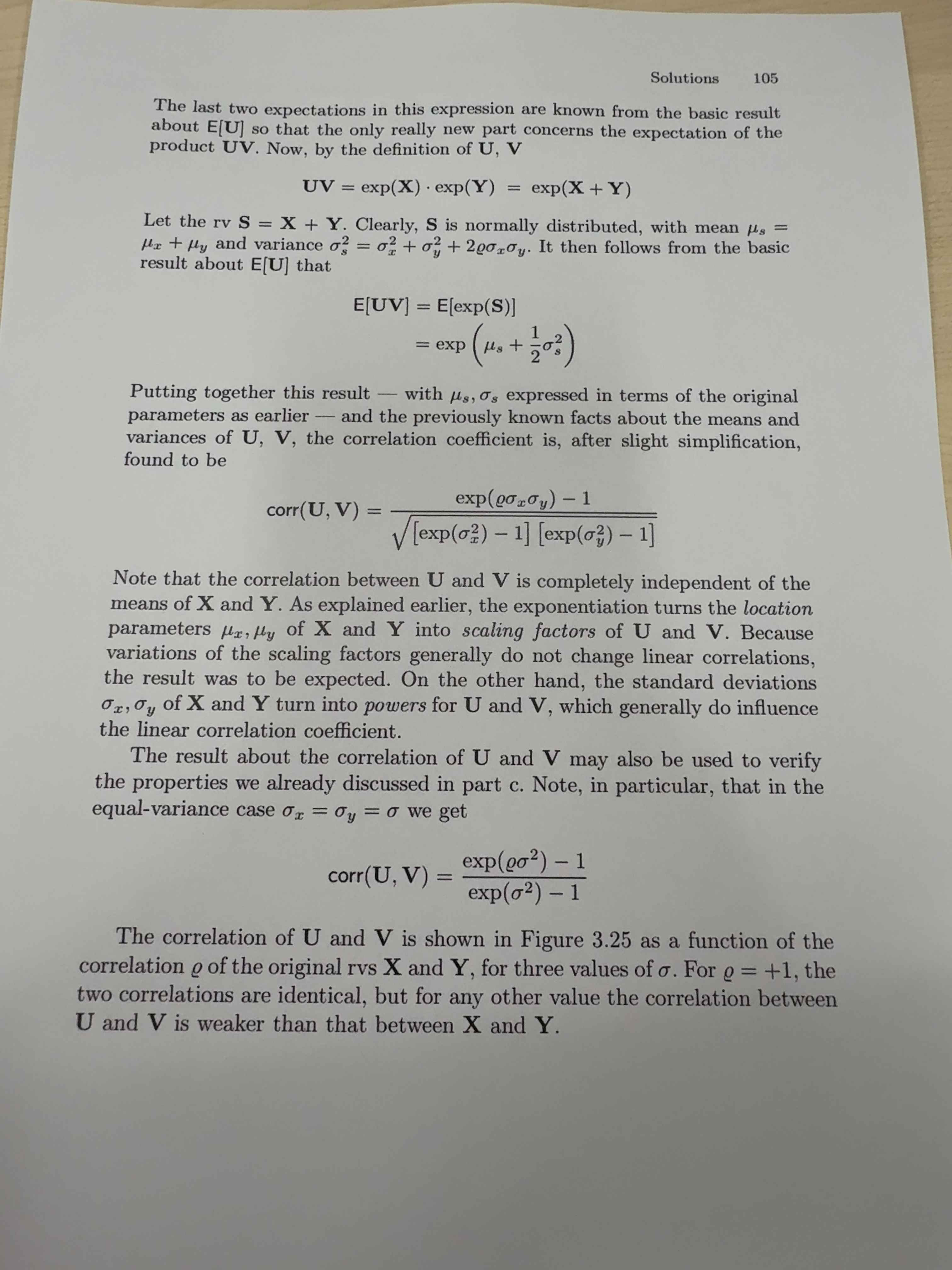} \\ \small{(a) Pitch}
\end{minipage}
\hfill
\begin{minipage}[t]{0.19\textwidth}
\centering
\includegraphics[width=\linewidth]{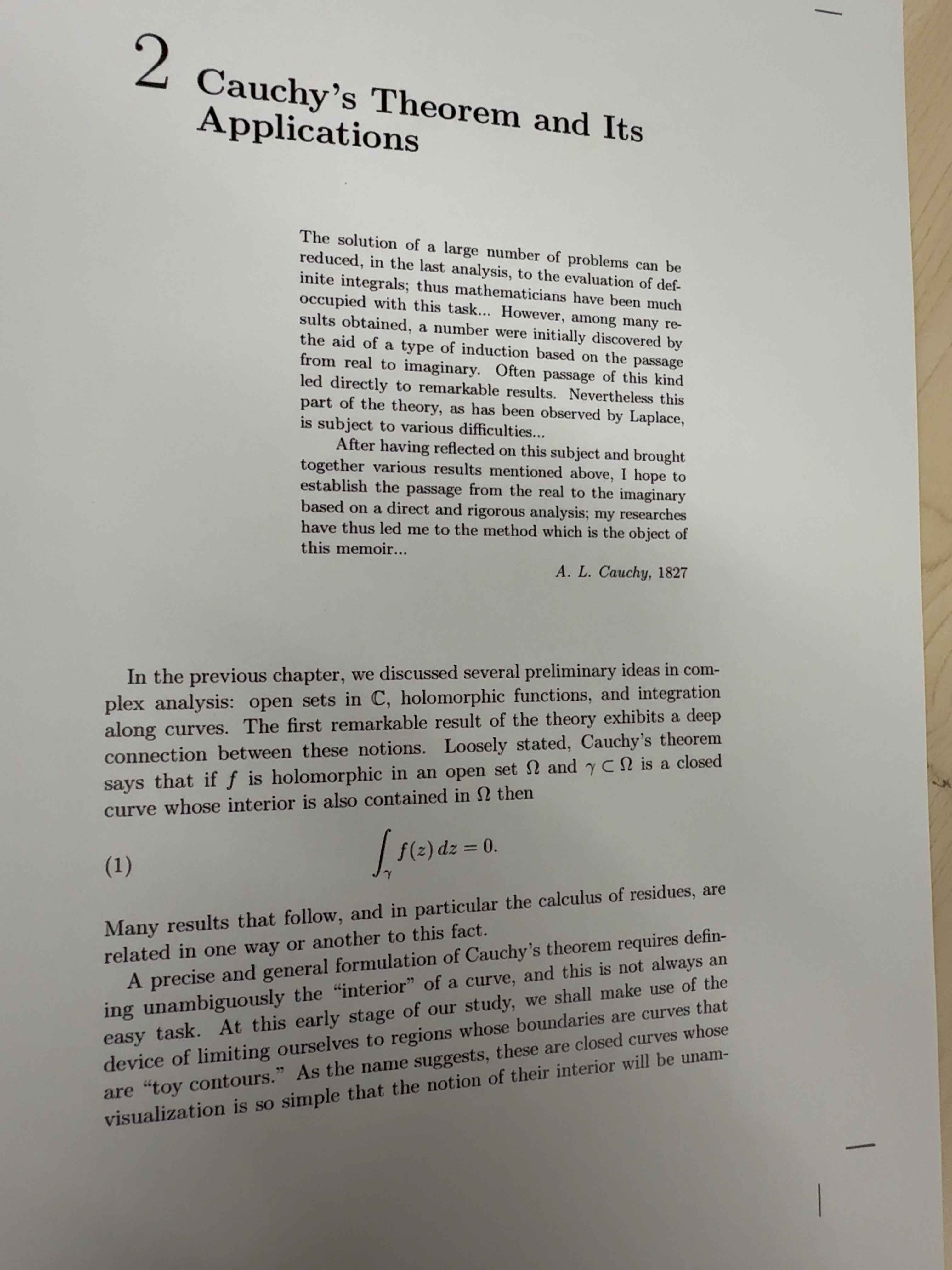} \\ \small{(b) Roll}
\end{minipage}
\hfill
\begin{minipage}[t]{0.19\textwidth}
\centering
\includegraphics[width=\linewidth]{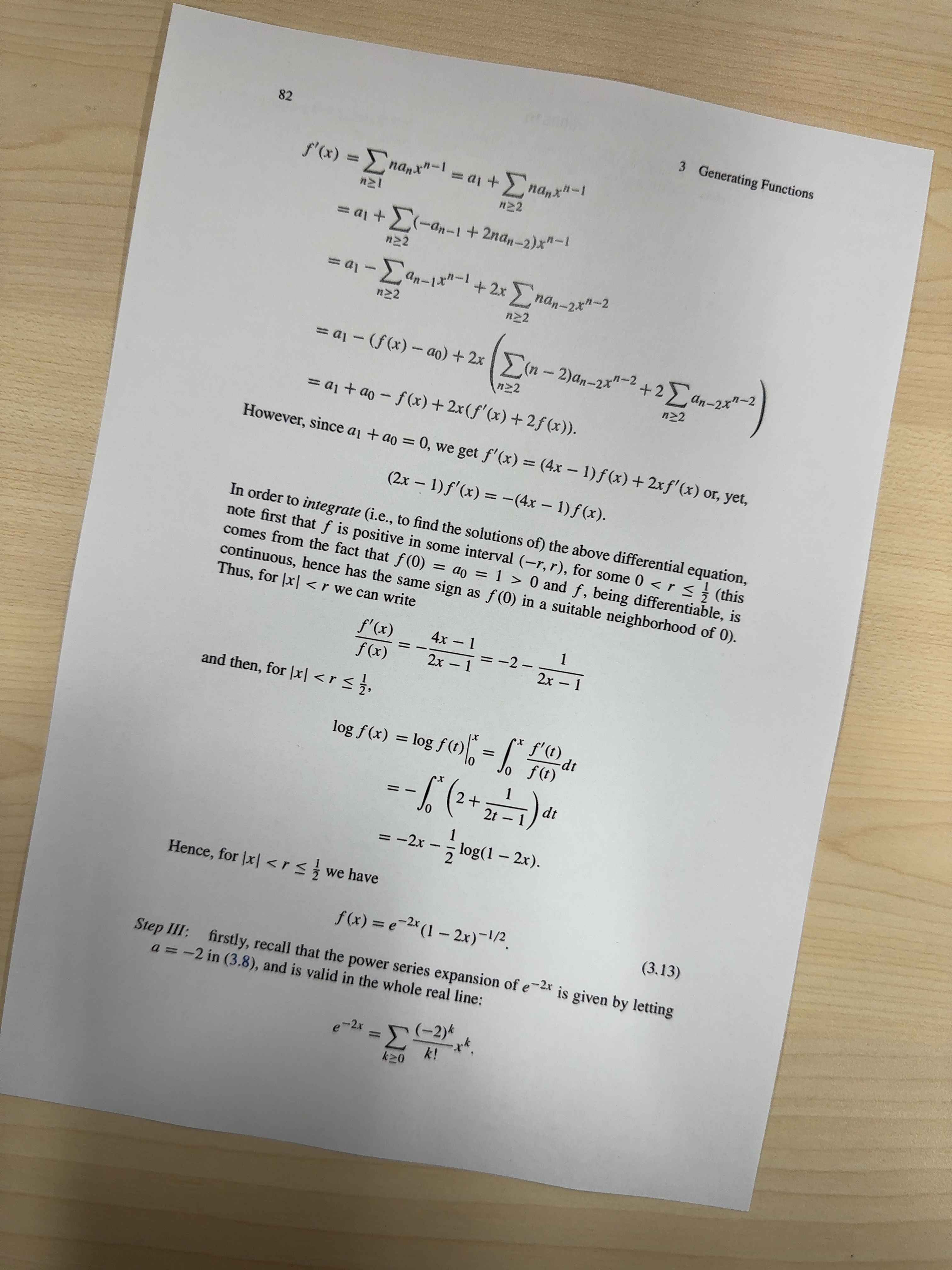} \\ \small{(c) Yaw}
\end{minipage}
\hfill
\begin{minipage}[t]{0.19\textwidth}
\centering
\includegraphics[width=\linewidth]{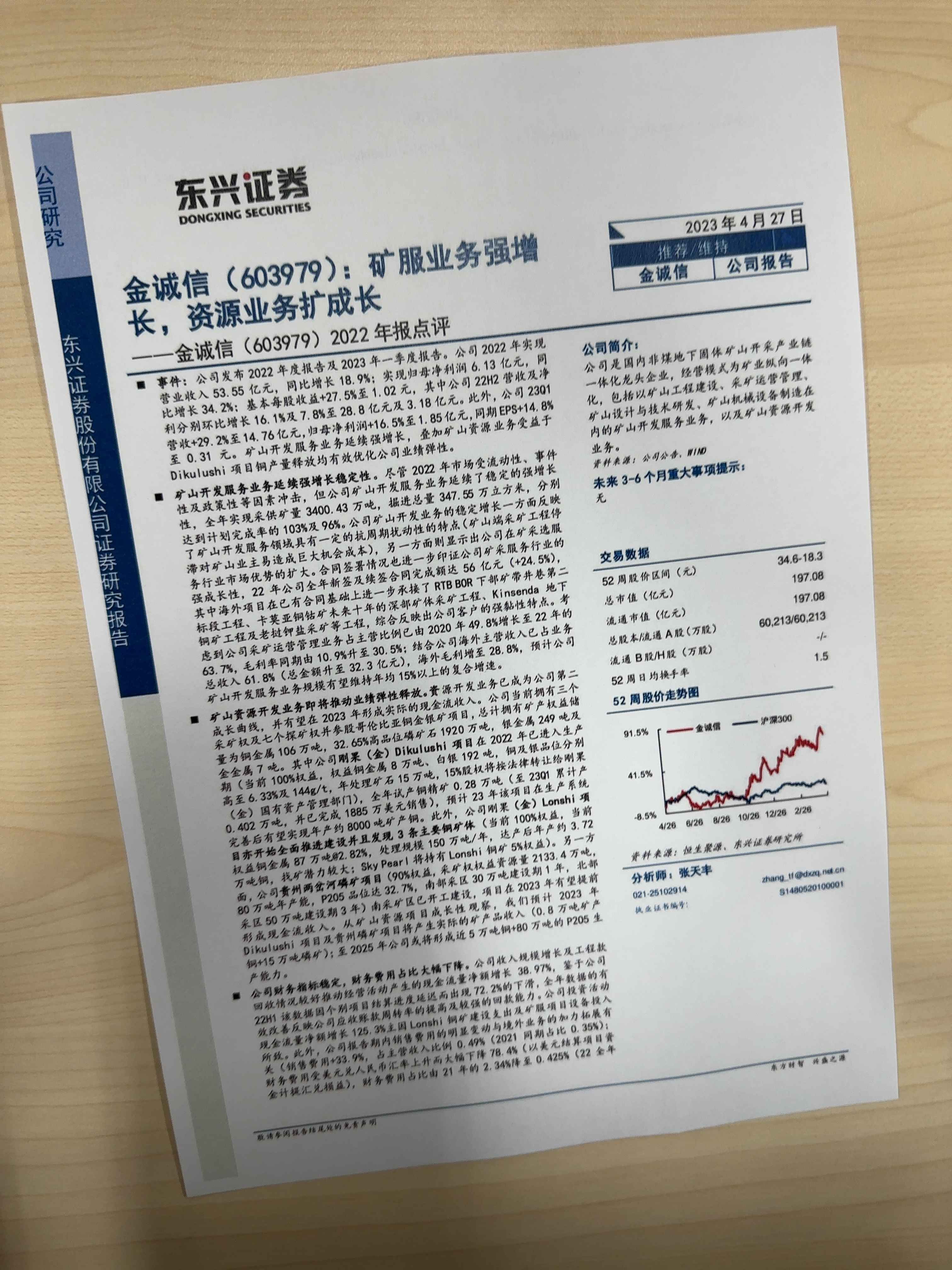} \\ \small{(d) Compound}
\end{minipage}
\hfill
\begin{minipage}[t]{0.19\textwidth}
\centering
\includegraphics[width=\linewidth]{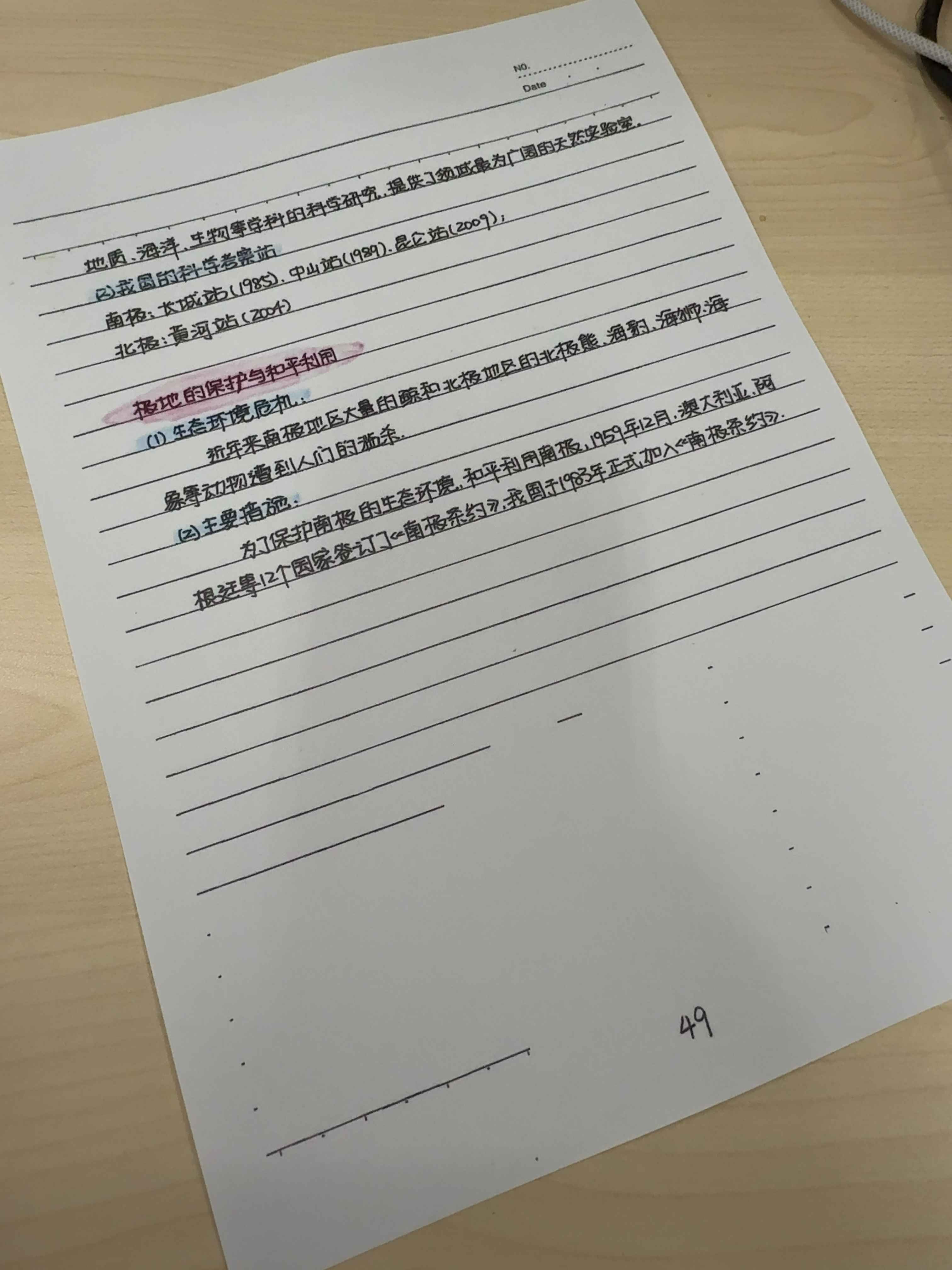} \\ \small{(e) Extreme}
\end{minipage}
\caption{Systematic 3D pose variations in the Skew scenario.}
\label{fig:skew_types}
\end{figure}

\paragraph{Skew.} 
The Skew scenario models 3D perspective distortions inherent in handheld photography. We design five pose configurations in Fig.~\ref{fig:skew_types}: \textit{Pitch} (a) causing vertical foreshortening; \textit{Roll} (b) producing horizontal trapezoidal distortion; \textit{Yaw} (c) resulting in planar rotation; \textit{Compound} (d) combining multi-axis rotations; and \textit{Extreme} (e) featuring aggressive tilt angles with severe edge distortions. These variations reflect the geometric challenges of real-world mobile document capture.

\subsection{Quality Audit and Refinement}
\label{subsec:quality}

To guarantee the reliability of Real5-OmniDocBench, we implement a multi-stage quality control pipeline combining automated anomaly detection with expert manual verification.

\paragraph{Machine-Assisted Anomaly Detection.}
We employ a committee of state-of-the-art (SOTA) VLMs to perform initial inference across all captured images. Samples exhibiting catastrophic performance drops (\eg, zero-score outliers) across the entire committee are flagged as potential anomalies. This phase efficiently locates samples suffering from severe hardware-induced failures, such as extreme motion blur, hardware-level overexposure, or accidental frame misalignment.

\paragraph{Expert Manual Verification and Recapture.}
All flagged samples undergo a rigorous manual audit. We perform three rounds of full-set inspection to ensure: (1) Semantic Correspondence: Each image correctly maps to its original digital counterpart in OmniDocBench v1.5; (2) Structural Integrity: No significant content is lost due to improper framing or occlusion; (3) Geometric Orthogonality: Image orientation is standardized, correcting any unintentional 90/180-degree rotations. Samples confirmed with irrecoverable artifacts (\eg, unreadable text due to severe defocus) are subjected to a recapture-and-replace loop until they meet our quality threshold.

\paragraph{Fidelity vs. Real-world Degradation.}
A critical distinction is made between unintentional capture errors and representative real-world degradation. While our 1200\,dpi printing ensures source fidelity, handheld capture inherently introduces a resolution gap compared to digital originals. We intentionally retain general clarity degradation as it reflects the objective constraints of mobile document parsing. Our audit protocol only filters out ``unusable'' samples (\eg, severe blur causing illegibility), while preserving authentic artifacts like moir\'{e} and minor noise. This ensures the benchmark evaluates the model's robustness to real-world deployment conditions rather than its tolerance for accidental photographic errors.

\paragraph{Dataset Completeness.}
Through continuous refinement, we maintain 100\% coverage of the original 1,355 OmniDocBench samples across all five scenarios. The final version of Real5-OmniDocBench represents the optimal convergence of physical realism and evaluation reliability, providing a stable foundation for benchmarking the next generation of VLMs.

\section{Evaluation Methodology and Metrics}
\label{sec:evaluation}

Real5-OmniDocBench is designed to be fully compatible with the evaluation ecosystem of OmniDocBench \cite{omnidocbench2024}, facilitating a rigorous cross-domain diagnostic analysis between pristine digital originals and their corresponding physical degradations.

\paragraph{Core Metrics and Scientific Logic.}
We adopt the multi-dimensional scoring framework of OmniDocBench, which decouples layout, content, and structural logic. The metrics are defined as follows:
\begin{itemize}
    \item \textbf{Text and Reading Order}: We utilize the Normalized Edit Distance (NED):
    $$NED = 1 - \frac{Lev(s_1, s_2)}{\max(|s_1|, |s_2|)}$$
    where $Lev$ denotes the Levenshtein distance. This is applied to both full-text OCR and reading order to measure character-level precision and topological consistency.
    \item \textbf{Formula Parsing (CDM)}: A key feature of this framework is the character detection matching (CDM) ~\cite{wang2025image} metric. Unlike string-based metrics, CDM converts formulas into computational graphs to evaluate structural equivalence, ensuring an objective assessment of mathematical logic even when physical noise induces minor syntactic variations.
    \item \textbf{Table Structure and Content (TEDS)}: We employ the Tree Edit Distance-based Similarity (TEDS) \cite{zhong2020image}:
    $$TEDS(G_a, G_p) = 1 - \frac{EditDist(G_a, G_p)}{\max(|G_a|, |G_p|)}$$
    where $G$ represents the table's tree structure. We report both \textbf{TEDS (Overall)} and TEDS-Struct to identify whether failures occur in content recognition or structural reconstruction.
    \item \textbf{Overall Metric}: The overall performance is quantified as follows:
    $$\text{Overall} = \frac{(1 - \text{Text NED}) \times 100 + \text{Table TEDS} + \text{Formula CDM}}{3}$$
\end{itemize}

\paragraph{Model Array and Alignment Protocol.}
To ensure the credibility of our benchmark, we categorize test subjects based on their alignment with official records:
\begin{itemize}
    \item \textbf{Officially Aligned Baselines}: We prioritize models verified in the original OmniDocBench report, including Qwen3-VL-235B,  Qwen2.5-VL-72B, Gemini-2.5 Pro, MinerU-1.8.2, and PP-StructureV3. Our evaluation principle follows a strict alignment-first protocol: we first perform inference on the official digital test set; only models that successfully reproduce official metrics are then evaluated on our Real5-OmniDocBench dataset. For models where official digital metrics cannot be currently aligned, we withhold their Real5-OmniDocBench scores to maintain evaluation rigor.
    \item \textbf{Supplemental Frontier Evaluations}: We additionally include the latest state-of-the-art models, such as Gemini-3 Pro, GPT-5.2, GLM-OCR, DeepSeek-OCR 2 and PaddleOCR-VL-1.6, using identical configurations to reflect the most recent progress in VLM robustness.
\end{itemize}

\section{Experiments and Analysis}
\label{sec:experiments}

\subsection{Main Results Overview}
\label{subsec:main_results}

Table~\ref{tab:main_results} presents a comprehensive performance analysis of various models on the Real5-OmniDocBench benchmark. The results indicate that different model architectures exhibit distinct robustness profiles when encountering physical degradations. Notably, the specialized PaddleOCR-VL-1.6 achieves an Overall score of 93.19, demonstrating high competitive efficiency at a 0.9B parameter scale compared to significantly larger counterparts, such as Qwen3-VL-235B (88.90) and Gemini-3 Pro (89.24). Across all five evaluated scenarios, this compact model maintains consistent performance, with metrics ranging from 91.25\% in Warping to 94.74\% in Scanning. Such stability across diverse physical conditions suggests that domain-specific architectural optimization and exposure to diverse artifacts during training may be more critical for real-world document parsing than raw parameter scaling, which primarily benefits semantic priors rather than geometric and optical invariance.

Within the General VLM category, Gemini-3 Pro shows consistent performance in maintaining semantic coherence. While Qwen3-VL-235B exhibits a specialized advantage in the Warping scenario (89.99\%), its performance fluctuates in Screen-Photography (89.27\%) and Skew (86.56\%) tasks. These fluctuations suggest that increasing parameter scale primarily enhances character recognition accuracy but may not inherently resolve geometric ambiguities introduced by 3D perspective transformations. In contrast, traditional Pipeline Tools show a higher sensitivity to environmental changes. For instance, both PP-StructureV3 and Marker-1.8.2 experience a notable decline in Skew scenarios compared to their Scanning results, primarily because heuristic-based layout analysis often struggles with large-angle perspective distortions.

\begin{table}[t]
\caption{Comprehensive evaluation on Real5-OmniDocBench. \textbf{S}: Scanning, \textbf{W}: Warping, \textbf{SP}: Screen-Photography, \textbf{I}: Illumination, \textbf{SK}: Skew. Best results are \textbf{bolded} and second-best are \underline{underlined}. }
\label{tab:main_results}
\centering
\resizebox{\textwidth}{!}{%
\renewcommand{\arraystretch}{1.1}
\begin{tabular}{llc|c|ccccc}
\toprule
\textbf{Model Type} & \textbf{Methods} & \textbf{Params} & \textbf{Overall$\uparrow$} & \textbf{S$\uparrow$} & \textbf{W$\uparrow$} & \textbf{SP$\uparrow$} & \textbf{I$\uparrow$} & \textbf{SK$\uparrow$}\\
\midrule
\multirow{2}{*}{Pipeline Tools} 
& Marker-1.8.2~\cite{vik2024marker} & - & 60.10 & 70.27 & 58.98 & 63.65 & 66.31 & 41.27 \\
& PP-StructureV3~\cite{cui2025paddleocr30technicalreport} & - & 64.45 & 84.68 & 59.34 & 66.89 & 73.38 & 37.98 \\
\midrule
\multirow{5}{*}{General VLMs} 
& GPT-5.2~\cite{gpt5_2} & - & 78.66 & 84.43 & 76.26 & 76.75 & 80.88 & 75.00 \\
& Qwen2.5-VL-72B~\cite{bai2025qwen25vltechnicalreport} & 72B & 86.92 & 86.19 & 87.77 & 86.48 & 87.25 & 86.90 \\
& Gemini-2.5 Pro~\cite{gemini25} & - & 88.21 & 89.25 & 87.63 & 87.11 & 87.97 & 89.07\\
& Qwen3-VL-235B~\cite{yang2025qwen3technicalreport} & 235B & 88.90 & 89.43 & 89.99 & 89.27 & 89.27 & 86.56 \\
& Kimi K2.5~\cite{kimiteam2026kimik25visualagentic} & 1T & 89.09 & 89.67 & 88.86 & 88.39 & 89.66 & 88.86 \\
& Gemini-3 Pro~\cite{gemini30} & - & 89.24 & 89.47 & 88.90 & 88.86 & 89.53 & 89.45 \\
\midrule
\multirow{8}{*}{Specialized VLMs}
& Deepseek-OCR 2~\cite{wei2026deepseekocr2visualcausal} & 3B & 73.01 & 89.59 & 66.53 & 71.65 & 76.02 & 61.28\\
& Deepseek-OCR~\cite{wei2025deepseekocrcontextsopticalcompression} & 3B & 73.99 & 86.17 & 67.20 & 75.31 & 78.10 & 63.01\\
& MinerU2-VLM~\cite{MinerU2} & 0.9B & 76.95 & 83.60 & 73.73 & 78.77 & 80.51 & 68.16\\
& MonkeyOCR-pro-3B~\cite{li2026monkeyocrdocumentparsingstructurerecognitionrelation} & 3.7B & 79.49 & 86.94 & 78.90 & 82.44 & 84.71 & 64.47\\
& Nanonets-OCR-s~\cite{Nanonets-OCR-S} & 3B & 84.19 & 85.52 & 83.56 & 84.86 & 85.01 & 81.98\\
& dots.ocr~\cite{li2025dotsocrmultilingualdocumentlayout} & 3B & 86.38 & 86.87 & 86.01 & 87.18 & 87.57 & 84.27\\
& PaddleOCR-VL~\cite{cui2025paddleocrvlboostingmultilingualdocument} & 0.9B & 85.54 & 92.11 & 85.97 & 82.54 & 89.61 & 77.47\\
& MinerU2.5~\cite{niu2025mineru25decoupledvisionlanguagemodel} & 1.2B & 85.61 & 90.06 & 83.76 & 89.41 & 89.57 & 75.24\\
& MinerU2.5-Pro~\cite{wang2026mineru2} & 1.2B & 88.96 & 92.11 & 88.72 & \underline{92.29} & 91.42 & 81.26\\
& GLM-OCR~\cite{glmocr} & 0.9B & 90.32 & 92.67 & 90.68 & 91.75 & 91.12 & 85.39\\
& PaddleOCR-VL-1.5~\cite{cui2026paddleocrvl15multitask09bvlm} & 0.9B & \underline{92.05} & \underline{93.43} & \underline{91.25} & 91.76 & \underline{92.16} & \underline{91.66}\\
& PaddleOCR-VL-1.6~\cite{zhang2026paddleocrvl16expandingfrontierdocument} & 0.9B & \textbf{93.19} & \textbf{94.74} & \textbf{92.48} & \textbf{92.78} & \textbf{93.28} & \textbf{92.66}\\
\bottomrule
\end{tabular}%
}
\end{table}

\subsection{Scenario-Level Diagnostic Analysis}
\label{subsec:scenario_analysis}

To provide a granular quantification of model robustness, we decompose the evaluation into four critical sub-tasks: Full-text OCR (Text$^{\text{E}}$), Formula Recognition (Formula$^{\text{CDM}}$), Table Reconstruction (Table$^{\text{TEDS}}$), and Reading Order consistency (RO$^{\text{E}}$). This multi-dimensional analysis helps identify the specific failure mechanisms of models under different physical artifacts, moving beyond simple accuracy metrics to understand structural and semantic fidelity.

\paragraph{Scanning \& Warping: Impact of Geometric Deformation on Structure.}
The Scanning scenario (Table~\ref{tab:scanning}) reflects the ideal performance bound, where documents are flat and evenly lit. Specialized VLMs achieve character-level precision nearly identical to digital-born benchmarks. However, in the Warping scenario (Table~\ref{tab:warping}), non-rigid deformations introduce complex curvilinear distortions that interfere with coordinate-based parsing. Our analysis indicates that while general VLMs maintain high recognition rates, their structural fidelity (TEDS) often declines more noticeably than that of PaddleOCR-VL-1.6. This suggests that paper curvature alters the spatial relationship between text lines, whereas models enhanced for geometric robustness can better anchor cell boundaries and formula components within distorted visual inputs.

\begin{table}[htbp]
    \centering
    \renewcommand{\arraystretch}{1.2}
    \setlength{\tabcolsep}{5pt}
    
    \begin{minipage}[t]{0.48\textwidth}
        \centering
        \caption{Sub-task: \textbf{Scanning}}
        \label{tab:scanning}
        \resizebox{\textwidth}{!}{%
        \begin{tabular}{lccccc}
        \toprule
        \textbf{Methods} & \textbf{Overall$\uparrow$} & \textbf{Text$^{\text{E}}\downarrow$} & \textbf{CDM$\uparrow$} & \textbf{TEDS$\uparrow$} & \textbf{RO$^{\text{E}}\downarrow$} \\
        \midrule
        PP-StructureV3~\cite{cui2025paddleocr30technicalreport} & 84.68 & 0.094 & 84.34 & 79.06 & 0.092 \\
        Marker-1.8.2~\cite{vik2024marker} & 70.27 & 0.223 & 77.03 & 56.05 & 0.238 \\
        \midrule
        Qwen3-VL-235B~\cite{yang2025qwen3technicalreport} & 89.43 & 0.059 & 89.01 & 85.19 & 0.066 \\
        Gemini-3 Pro~\cite{gemini30} & 89.47 & 0.071 & 88.16 & 87.37 & 0.078 \\
        \midrule
        MinerU2.5-Pro~\cite{wang2026mineru2} & 92.11 & \underline{0.040} & 89.77 & 90.57 & \textbf{0.043} \\
        GLM-OCR~\cite{glmocr} & \underline{92.67} & 0.054 & \underline{91.10} & \underline{92.28} & \underline{0.061} \\
        PaddleOCR-VL-1.6~\cite{zhang2026paddleocrvl16expandingfrontierdocument} & \textbf{94.74} & \textbf{0.036} & \textbf{93.65} & \textbf{94.12} & \textbf{0.043} \\
        \bottomrule
        \end{tabular}}
    \end{minipage}
    \hfill
    \begin{minipage}[t]{0.48\textwidth}
        \centering
        \caption{Sub-task: \textbf{Warping}}
        \label{tab:warping}
        \resizebox{\textwidth}{!}{%
        \begin{tabular}{lccccc}
        \toprule
        \textbf{Methods} & \textbf{Overall$\uparrow$} & \textbf{Text$^{\text{E}}\downarrow$} & \textbf{CDM$\uparrow$} & \textbf{TEDS$\uparrow$} & \textbf{RO$^{\text{E}}\downarrow$} \\
        \midrule
        PP-StructureV3~\cite{cui2025paddleocr30technicalreport} & 59.34 & 0.376 & 68.22 & 47.40 & 0.261 \\
        Marker-1.8.2~\cite{vik2024marker} & 58.98 & 0.349 & 72.71 & 39.08 & 0.390 \\
        \midrule
        Qwen3-VL-235B~\cite{yang2025qwen3technicalreport} & 89.99 & \underline{0.051} & 89.06 & 85.95 & \underline{0.064} \\
        Gemini-3 Pro~\cite{gemini30} & 88.90 & 0.086 & 88.10 & 87.20 & 0.087 \\
        \midrule
        MinerU2.5-Pro~\cite{wang2026mineru2} & 88.72 & 0.100 & 87.81 & 88.36 & 0.076 \\
        GLM-OCR~\cite{glmocr} & \underline{90.68} & 0.071 & \underline{90.30} & \underline{88.78} & 0.100 \\
        PaddleOCR-VL-1.6~\cite{zhang2026paddleocrvl16expandingfrontierdocument} & \textbf{92.48} & \textbf{0.049} & \textbf{91.63} & \textbf{90.66} & \textbf{0.061} \\
        \bottomrule
        \end{tabular}}
    \end{minipage}
\end{table}

\paragraph{Screen-Photography \& Illumination: Optical Artifacts and Consistency.}
Optical interference remains a significant bottleneck for real-world document intelligence. Moir\'{e} patterns in Screen-Photography (Table~\ref{tab:screen}) and localized overexposure in Illumination (Table~\ref{tab:illumination}) often trigger "layout fragmentation" in traditional systems. Our experiments show that pipeline-based tools frequently misinterpret reflections as structural boundaries, leading to a sharp decrease in Reading Order (RO$^{\text{E}}$) scores. In contrast, PaddleOCR-VL-1.6 demonstrates superior visual consistency. By integrating diverse lighting-augmented data during multi-task training, the model learns to utilize global contextual cues to maintain parsing precision even when local visual contrast is compromised.

\begin{table}[htbp]
    \centering
    \renewcommand{\arraystretch}{1.2}
    \begin{minipage}[t]{0.48\textwidth}
        \centering
        \caption{Sub-task: \textbf{Screen-Photo}}
        \label{tab:screen}
        \resizebox{\textwidth}{!}{%
        \begin{tabular}{lccccc}
        \toprule
        \textbf{Methods} & \textbf{Overall$\uparrow$} & \textbf{Text$^{\text{E}}\downarrow$} & \textbf{CDM$\uparrow$} & \textbf{TEDS$\uparrow$} & \textbf{RO$^{\text{E}}\downarrow$} \\
        \midrule
        PP-StructureV3~\cite{cui2025paddleocr30technicalreport} & 66.89 & 0.204 & 73.26 & 47.82 & 0.165 \\
        Marker-1.8.2~\cite{vik2024marker}   & 63.65 & 0.290 & 72.73 & 47.21 & 0.325 \\
        \midrule
        Qwen3-VL-235B~\cite{yang2025qwen3technicalreport}  & 89.27 & 0.068 & 88.72 & 85.85 & 0.071 \\
        Gemini-3 Pro~\cite{gemini30} & 88.86 & 0.084 & 87.33 & 87.65 & 0.087 \\
        \midrule
        MinerU2.5-Pro~\cite{wang2026mineru2}      & 91.29 & \underline{0.050} & 87.41 & 91.44 & \textbf{0.044} \\
        GLM-OCR~\cite{glmocr} & \underline{91.75} & 0.063 & \underline{89.83} & \underline{91.66} & 0.071 \\
        PaddleOCR-VL-1.6~\cite{zhang2026paddleocrvl16expandingfrontierdocument} & \textbf{92.78} & \textbf{0.045} & \textbf{90.64} & \textbf{92.19} & \underline{0.054} \\
        \bottomrule
        \end{tabular}}
    \end{minipage}
    \hfill
    \begin{minipage}[t]{0.48\textwidth}
        \centering
        \caption{Sub-task: \textbf{Illumination}}
        \label{tab:illumination}
        \resizebox{\textwidth}{!}{%
        \begin{tabular}{lccccc}
        \toprule
        \textbf{Methods} & \textbf{Overall$\uparrow$} & \textbf{Text$^{\text{E}}\downarrow$} & \textbf{CDM$\uparrow$} & \textbf{TEDS$\uparrow$} & \textbf{RO$^{\text{E}}\downarrow$} \\
        \midrule
        PP-StructureV3~\cite{cui2025paddleocr30technicalreport} & 73.38 & 0.158 & 77.75 & 58.19 & 0.126 \\
        Marker-1.8.2~\cite{vik2024marker}   & 66.31 & 0.259 & 74.80 & 50.03 & 0.337 \\
        \midrule
        Qwen3-VL-235B~\cite{yang2025qwen3technicalreport}  & 89.27 & 0.060 & 87.81 & 86.05 & 0.070 \\
        Gemini-3 Pro~\cite{gemini30} & 89.53 & 0.073 & 87.78 & 88.14 & 0.080 \\
        \midrule
        MinerU2.5-Pro~\cite{wang2026mineru2}      & \underline{91.31} & \underline{0.050} & 88.15 & \underline{90.76} & \underline{0.053} \\
        GLM-OCR~\cite{glmocr} & 91.12 & 0.059 & \underline{91.02} & 88.20 & 0.071 \\
        PaddleOCR-VL-1.6~\cite{zhang2026paddleocrvl16expandingfrontierdocument} & \textbf{93.28} & \textbf{0.042} & \textbf{92.61} & \textbf{91.48} & \textbf{0.051} \\
        \bottomrule
        \end{tabular}}
    \end{minipage}
\end{table}

\paragraph{Skew: Perspective Correction and Reading Order Recovery.}
The Skew scenario (Table~\ref{tab:skew}) represents one of the most demanding tasks in our benchmark, testing the limits of 3D-to-2D projection recovery. Large-angle perspective tilting induces a "line compression" effect, where text lines that are originally parallel appear to converge or overlap. Traditional systems relying on axis-aligned projections are prone to severe sequence errors. Experimental data show that PaddleOCR-VL-1.6 maintains a robust RO$^{\text{E}}$ of 0.058, outperforming several large-scale general models. This performance suggests the model has internalized a flexible mapping of perspective cues, allowing it to accurately reconstruct the logical flow of a document even when the visual layout is aggressively skewed.

\begin{table}[htbp]
    \centering
    \caption{Detailed performance decomposition on the \textbf{Skew} scenario.}
    \label{tab:skew}
    \vspace{2mm}
    \renewcommand{\arraystretch}{1.2}
    \resizebox{0.8\textwidth}{!}{%
    \begin{tabular}{lccccc}
    \toprule
    \textbf{Methods} & \textbf{Overall$\uparrow$} & \textbf{Text$^{\text{E}}\downarrow$} & \textbf{CDM$\uparrow$} & \textbf{TEDS$\uparrow$} & \textbf{RO$^{\text{E}}\downarrow$} \\
    \midrule
    PP-StructureV3~\cite{cui2025paddleocr30technicalreport} & 37.98 & 0.557 & 44.37 & 25.27 & 0.417 \\
    Marker-1.8.2~\cite{vik2024marker} & 41.27 & 0.536 & 60.16 & 17.23 & 0.543 \\
    \midrule
    Qwen3-VL-235B~\cite{yang2025qwen3technicalreport} & 86.56 & \underline{0.077} & 83.96 & 83.41 & \underline{0.091} \\
    Gemini-3 Pro~\cite{gemini30} & \underline{89.45} & 0.080 & \underline{88.33} & \underline{88.06} & 0.092 \\
    \midrule
    MinerU2.5-Pro~\cite{wang2026mineru2} & 81.26 & 0.212 & 83.92 & 81.07 & 0.107 \\
    GLM-OCR~\cite{glmocr} & 85.39 & 0.099 & 85.78 & 80.28 & 0.156 \\
    PaddleOCR-VL-1.6~\cite{zhang2026paddleocrvl16expandingfrontierdocument} & \textbf{92.66} & \textbf{0.045} & \textbf{91.44} & \textbf{91.04} & \textbf{0.058} \\
    \bottomrule
    \end{tabular}}
\end{table}

\section{Discussion}
\label{sec:discussion}

\paragraph{Insights on Domain Gap and Robustness.} 
Systematic evaluation on Real5-Omni-DocBench reveals a pronounced performance degradation when models transition from digital-born environments to physical-world captures. Results indicate that while large-scale General VLMs possess strong semantic priors, their zero-shot transferability to non-rigid geometric transformations (\eg, Warping) is not strictly correlated with parameter scale. This suggests the current bottleneck lies less in linguistic modeling and more in visual feature alignment under stochastic distortions. The persistence of this "reality gap" emphasizes the need for benchmarks quantifying robustness beyond standard accuracy, focusing on the intersection of 3D vision and document analysis.

\paragraph{Efficacy of Specialized vs. General Architectures.} 
Our results provide a nuanced perspective on architectural choices. Although General VLMs demonstrate superior adaptability in open-domain coherence, specialized models optimized for multi-task parsing often exhibit higher structural fidelity in constrained resource settings. The competitive performance of compact architectures suggests that domain-specific fine-tuning on high-quality augmented data can effectively compensate for smaller parameter counts. Conversely, failures in traditional Pipeline Tools under Skew and Illumination highlight the limitations of heuristic systems, advocating for unified end-to-end architectures that jointly optimize optical correction and structural parsing.

\section{Conclusion}
\label{sec:conclusion}

In this paper, we introduced Real5-OmniDocBench, a comprehensive benchmark evaluating document parsing robustness across five real-world scenarios. By reconstructing 1,355 high-fidelity images based on the OmniDocBench v1.5 corpus, we provide a standardized platform for assessing how physical artifacts impact model performance. 

Our benchmarking reveals significant vulnerabilities in current parsing technologies, particularly regarding complex 3D distortions and localized illumination variations. Results highlight that while compact, specialized models achieve high efficiency, there remains a substantial margin for universal robustness. We hope Real5-OmniDocBench will catalyze research toward more resilient models capable of handling the inherent complexities of practical applications.

\bibliography{main}

@misc{omnidocbench2024,
      title={OmniDocBench: Benchmarking Diverse PDF Document Parsing with Comprehensive Annotations}, 
      author={Linke Ouyang and Yuan Qu and Hongbin Zhou and Jiawei Zhu and Rui Zhang and Qunshu Lin and Bin Wang and Zhiyuan Zhao and Man Jiang and Xiaomeng Zhao and Jin Shi and Fan Wu and Pei Chu and Minghao Liu and Zhenxiang Li and Chao Xu and Bo Zhang and Botian Shi and Zhongying Tu and Conghui He},
      year={2024},
      eprint={2412.07626},
      archivePrefix={arXiv},
      primaryClass={cs.CV},
      url={https://arxiv.org/abs/2412.07626}, 
}

@misc{olmocrbench2025,
      title={{olmOCR: Unlocking Trillions of Tokens in PDFs with Vision Language Models}},
      author={Jake Poznanski and Jon Borchardt and Jason Dunkelberger and Regan Huff and Daniel Lin and Aman Rangapur and Christopher Wilhelm and Kyle Lo and Luca Soldaini},
      year={2025},
      eprint={2502.18443},
      archivePrefix={arXiv},
      primaryClass={cs.CL},
      url={https://arxiv.org/abs/2502.18443},
}

@misc{docptbench2025,
  title={DocPTBench: Benchmarking End-to-End Photographed Document Parsing and Translation},
  author={Yongkun Du and Pinxuan Chen and Xuye Ying and Zhineng Chen},
  year={2025},
  eprint={2511.18434},
  archivePrefix={arXiv},
  primaryClass={cs.CV},
  url={https://arxiv.org/abs/2511.18434}
}

@inproceedings{icdar2019,
  title={Icdar2019 competition on scanned receipt ocr and information extraction},
  author={Huang, Zheng and Chen, Kai and He, Jianhua and Bai, Xiang and Karatzas, Dimosthenis and Lu, Shijian and Jawahar, CV},
  booktitle={2019 International Conference on Document Analysis and Recognition (ICDAR)},
  pages={1516--1520},
  year={2019},
  organization={IEEE}
}

@inproceedings{docres2024,
  title={Docres: A generalist model toward unifying document image restoration tasks},
  author={Zhang, Jiaxin and Peng, Dezhi and Liu, Chongyu and Zhang, Peirong and Jin, Lianwen},
  booktitle={Proceedings of the IEEE/CVF Conference on Computer Vision and Pattern Recognition},
  pages={15654--15664},
  year={2024}
}

@misc{fu2025ocrbenchv2improvedbenchmark,
      title={OCRBench v2: An Improved Benchmark for Evaluating Large Multimodal Models on Visual Text Localization and Reasoning}, 
      author={Ling Fu and Zhebin Kuang and Jiajun Song and Mingxin Huang and Biao Yang and Yuzhe Li and Linghao Zhu and Qidi Luo and Xinyu Wang and Hao Lu and Zhang Li and Guozhi Tang and Bin Shan and Chunhui Lin and Qi Liu and Binghong Wu and Hao Feng and Hao Liu and Can Huang and Jingqun Tang and Wei Chen and Lianwen Jin and Yuliang Liu and Xiang Bai},
      year={2025},
      eprint={2501.00321},
      archivePrefix={arXiv},
      primaryClass={cs.CV},
      url={https://arxiv.org/abs/2501.00321}, 
}

@misc{bai2025qwen3vltechnicalreport,
      title={Qwen3-VL Technical Report}, 
      author={Shuai Bai and Yuxuan Cai and Ruizhe Chen and Keqin Chen and Xionghui Chen and Zesen Cheng and Lianghao Deng and Wei Ding and Chang Gao and Chunjiang Ge and Wenbin Ge and Zhifang Guo and Qidong Huang and Jie Huang and Fei Huang and Binyuan Hui and Shutong Jiang and Zhaohai Li and Mingsheng Li and Mei Li and Kaixin Li and Zicheng Lin and Junyang Lin and Xuejing Liu and Jiawei Liu and Chenglong Liu and Yang Liu and Dayiheng Liu and Shixuan Liu and Dunjie Lu and Ruilin Luo and Chenxu Lv and Rui Men and Lingchen Meng and Xuancheng Ren and Xingzhang Ren and Sibo Song and Yuchong Sun and Jun Tang and Jianhong Tu and Jianqiang Wan and Peng Wang and Pengfei Wang and Qiuyue Wang and Yuxuan Wang and Tianbao Xie and Yiheng Xu and Haiyang Xu and Jin Xu and Zhibo Yang and Mingkun Yang and Jianxin Yang and An Yang and Bowen Yu and Fei Zhang and Hang Zhang and Xi Zhang and Bo Zheng and Humen Zhong and Jingren Zhou and Fan Zhou and Jing Zhou and Yuanzhi Zhu and Ke Zhu},
      year={2025},
      eprint={2511.21631},
      archivePrefix={arXiv},
      primaryClass={cs.CV},
      url={https://arxiv.org/abs/2511.21631}, 
}

@misc{vik2024marker,
  author       = {Vik Paruchuri},
  title        = {Marker},
  year         = {2025},
  howpublished = {\url{https://github.com/datalab-to/marker}},
  note         = {Accessed: 2025-09-25},
}

@misc{cui2025paddleocr30technicalreport,
      title={PaddleOCR 3.0 Technical Report}, 
      author={Cheng Cui and Ting Sun and Manhui Lin and Tingquan Gao and Yubo Zhang and Jiaxuan Liu and Xueqing Wang and Zelun Zhang and Changda Zhou and Hongen Liu and Yue Zhang and Wenyu Lv and Kui Huang and Yichao Zhang and Jing Zhang and Jun Zhang and Yi Liu and Dianhai Yu and Yanjun Ma},
      year={2025},
      eprint={2507.05595},
      archivePrefix={arXiv},
      primaryClass={cs.CV},
      url={https://arxiv.org/abs/2507.05595}, 
}

@misc{gpt5_2,
  title = {GPT-5.2 System Card},
  url = {https://cdn.openai.com/pdf/3a4153c8-c748-4b71-8e31-aecbde944f8d/oai_5_2_system-card.pdf},
  author = {OpenAI.},
  year = {2025}
}

@misc{bai2025qwen25vltechnicalreport,
      title={Qwen2.5-VL Technical Report}, 
      author={Shuai Bai and Keqin Chen and Xuejing Liu and Jialin Wang and Wenbin Ge and Sibo Song and Kai Dang and Peng Wang and Shijie Wang and Jun Tang and Humen Zhong and Yuanzhi Zhu and Mingkun Yang and Zhaohai Li and Jianqiang Wan and Pengfei Wang and Wei Ding and Zheren Fu and Yiheng Xu and Jiabo Ye and Xi Zhang and Tianbao Xie and Zesen Cheng and Hang Zhang and Zhibo Yang and Haiyang Xu and Junyang Lin},
      year={2025},
      eprint={2502.13923},
      archivePrefix={arXiv},
      primaryClass={cs.CV},
      url={https://arxiv.org/abs/2502.13923}, 
}

@misc{gemini25,
  author={{Google DeepMind}},
  title={Gemini 2.5},
  howpublished={\url{https://blog.google/technology/google-deepmind/gemini-model-thinking-updates-march-2025/}},
  year={2025}
}

@misc{yang2025qwen3technicalreport,
      title={Qwen3 Technical Report}, 
      author={An Yang and Anfeng Li and Baosong Yang and Beichen Zhang and Binyuan Hui and Bo Zheng and Bowen Yu and Chang Gao and Chengen Huang and Chenxu Lv and Chujie Zheng and Dayiheng Liu and Fan Zhou and Fei Huang and Feng Hu and Hao Ge and Haoran Wei and Huan Lin and Jialong Tang and Jian Yang and Jianhong Tu and Jianwei Zhang and Jianxin Yang and Jiaxi Yang and Jing Zhou and Jingren Zhou and Junyang Lin and Kai Dang and Keqin Bao and Kexin Yang and Le Yu and Lianghao Deng and Mei Li and Mingfeng Xue and Mingze Li and Pei Zhang and Peng Wang and Qin Zhu and Rui Men and Ruize Gao and Shixuan Liu and Shuang Luo and Tianhao Li and Tianyi Tang and Wenbiao Yin and Xingzhang Ren and Xinyu Wang and Xinyu Zhang and Xuancheng Ren and Yang Fan and Yang Su and Yichang Zhang and Yinger Zhang and Yu Wan and Yuqiong Liu and Zekun Wang and Zeyu Cui and Zhenru Zhang and Zhipeng Zhou and Zihan Qiu},
      year={2025},
      eprint={2505.09388},
      archivePrefix={arXiv},
      primaryClass={cs.CL},
      url={https://arxiv.org/abs/2505.09388}, 
}

@misc{gemini30,
  author={{Google DeepMind}},
  title={Gemini 3.0},
  howpublished={\url{https://blog.google/products-and-platforms/products/gemini/gemini-3-collection/}},
  year={2025}
}

@misc{wei2025deepseekocrcontextsopticalcompression,
      title={DeepSeek-OCR: Contexts Optical Compression}, 
      author={Haoran Wei and Yaofeng Sun and Yukun Li},
      year={2025},
      eprint={2510.18234},
      archivePrefix={arXiv},
      primaryClass={cs.CV},
      url={https://arxiv.org/abs/2510.18234}, 
}

@misc{MinerU2,
  title={MinerU2.0-2505-0.9B},
  howpublished={\url{https://huggingface.co/opendatalab/MinerU2.0-2505-0.9B}},
  author={{opendatalab}},
  year={2025},
}

@misc{li2026monkeyocrdocumentparsingstructurerecognitionrelation,
      title={MonkeyOCR: Document Parsing with a Structure-Recognition-Relation Triplet Paradigm}, 
      author={Zhang Li and Yuliang Liu and Qiang Liu and Zhiyin Ma and Ziyang Zhang and Shuo Zhang and Biao Yang and Zidun Guo and Jiarui Zhang and Xinyu Wang and Xiang Bai},
      year={2026},
      eprint={2506.05218},
      archivePrefix={arXiv},
      primaryClass={cs.CV},
      url={https://arxiv.org/abs/2506.05218}, 
}

@misc{Nanonets-OCR-S,
  title={Nanonets-OCR-S: A model for transforming documents into structured markdown with intelligent content recognition and semantic tagging},
  author={Souvik Mandal and Ashish Talewar and Paras Ahuja and Prathamesh Juvatkar},
  year={2025},
}

@misc{cui2025paddleocrvlboostingmultilingualdocument,
      title={PaddleOCR-VL: Boosting Multilingual Document Parsing via a 0.9B Ultra-Compact Vision-Language Model}, 
      author={Cheng Cui and Ting Sun and Suyin Liang and Tingquan Gao and Zelun Zhang and Jiaxuan Liu and Xueqing Wang and Changda Zhou and Hongen Liu and Manhui Lin and Yue Zhang and Yubo Zhang and Handong Zheng and Jing Zhang and Jun Zhang and Yi Liu and Dianhai Yu and Yanjun Ma},
      year={2025},
      eprint={2510.14528},
      archivePrefix={arXiv},
      primaryClass={cs.CV},
      url={https://arxiv.org/abs/2510.14528}, 
}

@misc{niu2025mineru25decoupledvisionlanguagemodel,
      title={MinerU2.5: A Decoupled Vision-Language Model for Efficient High-Resolution Document Parsing}, 
      author={Junbo Niu and Zheng Liu and Zhuangcheng Gu and Bin Wang and Linke Ouyang and Zhiyuan Zhao and Tao Chu and Tianyao He and Fan Wu and Qintong Zhang and Zhenjiang Jin and Guang Liang and Rui Zhang and Wenzheng Zhang and Yuan Qu and Zhifei Ren and Yuefeng Sun and Yuanhong Zheng and Dongsheng Ma and Zirui Tang and Boyu Niu and Ziyang Miao and Hejun Dong and Siyi Qian and Junyuan Zhang and Jingzhou Chen and Fangdong Wang and Xiaomeng Zhao and Liqun Wei and Wei Li and Shasha Wang and Ruiliang Xu and Yuanyuan Cao and Lu Chen and Qianqian Wu and Huaiyu Gu and Lindong Lu and Keming Wang and Dechen Lin and Guanlin Shen and Xuanhe Zhou and Linfeng Zhang and Yuhang Zang and Xiaoyi Dong and Jiaqi Wang and Bo Zhang and Lei Bai and Pei Chu and Weijia Li and Jiang Wu and Lijun Wu and Zhenxiang Li and Guangyu Wang and Zhongying Tu and Chao Xu and Kai Chen and Yu Qiao and Bowen Zhou and Dahua Lin and Wentao Zhang and Conghui He},
      year={2025},
      eprint={2509.22186},
      archivePrefix={arXiv},
      primaryClass={cs.CV},
      url={https://arxiv.org/abs/2509.22186}, 
}

@misc{li2025dotsocrmultilingualdocumentlayout,
      title={dots.ocr: Multilingual Document Layout Parsing in a Single Vision-Language Model}, 
      author={Yumeng Li and Guang Yang and Hao Liu and Bowen Wang and Colin Zhang},
      year={2025},
      eprint={2512.02498},
      archivePrefix={arXiv},
      primaryClass={cs.CV},
      url={https://arxiv.org/abs/2512.02498}, 
}

@misc{cui2026paddleocrvl15multitask09bvlm,
      title={PaddleOCR-VL-1.5: Towards a Multi-Task 0.9B VLM for Robust In-the-Wild Document Parsing}, 
      author={Cheng Cui and Ting Sun and Suyin Liang and Tingquan Gao and Zelun Zhang and Jiaxuan Liu and Xueqing Wang and Changda Zhou and Hongen Liu and Manhui Lin and Yue Zhang and Yubo Zhang and Yi Liu and Dianhai Yu and Yanjun Ma},
      year={2026},
      eprint={2601.21957},
      archivePrefix={arXiv},
      primaryClass={cs.CV},
      url={https://arxiv.org/abs/2601.21957}, 
}

@misc{wilddoc2025,
      title={WildDoc: How Far Are We from Achieving Comprehensive and Robust Document Understanding in the Wild?}, 
      author={An-Lan Wang and Jingqun Tang and Liao Lei and Hao Feng and Qi Liu and Xiang Fei and Jinghui Lu and Han Wang and Weiwei Liu and Hao Liu and Yuliang Liu and Xiang Bai and Can Huang},
      year={2025},
      eprint={2505.11015},
      archivePrefix={arXiv},
      primaryClass={cs.CV},
      url={https://arxiv.org/abs/2505.11015}, 
}

@misc{chen2025oceanocrgeneralocrapplication,
      title={Ocean-OCR: Towards General OCR Application via a Vision-Language Model}, 
      author={Song Chen and Xinyu Guo and Yadong Li and Tao Zhang and Mingan Lin and Dongdong Kuang and Youwei Zhang and Lingfeng Ming and Fengyu Zhang and Yuran Wang and Jianhua Xu and Zenan Zhou and Weipeng Chen},
      year={2025},
      eprint={2501.15558},
      archivePrefix={arXiv},
      primaryClass={cs.CV},
      url={https://arxiv.org/abs/2501.15558}, 
}

@inproceedings{wang2025image,
  title={Image over text: Transforming formula recognition evaluation with character detection matching},
  author={Wang, Bin and Wu, Fan and Ouyang, Linke and Gu, Zhuangcheng and Zhang, Rui and Xia, Renqiu and Shi, Botian and Zhang, Bo and He, Conghui},
  booktitle={Proceedings of the Computer Vision and Pattern Recognition Conference},
  pages={19681--19690},
  year={2025}
}

@inproceedings{zhong2020image,
  title={Image-based table recognition: data, model, and evaluation},
  author={Zhong, Xu and ShafieiBavani, Elaheh and Jimeno Yepes, Antonio},
  booktitle={European conference on computer vision},
  pages={564--580},
  year={2020},
  organization={Springer}
}

@misc{wei2026deepseekocr2visualcausal,
      title={DeepSeek-OCR 2: Visual Causal Flow}, 
      author={Haoran Wei and Yaofeng Sun and Yukun Li},
      year={2026},
      eprint={2601.20552},
      archivePrefix={arXiv},
      primaryClass={cs.CV},
      url={https://arxiv.org/abs/2601.20552}, 
}

@misc{glmocr,
  title={GLM-OCR},
  howpublished={\url{https://huggingface.co/zai-org/GLM-OCR}},
  author={{zai-org}},
  year={2026},
}

@misc{zhang2026paddleocrvl16expandingfrontierdocument,
      title={PaddleOCR-VL-1.6: Expanding the Frontier of Document Parsing with Under-Optimized Region Refinement and Progressive Post-Training}, 
      author={Zelun Zhang and Hongen Liu and Suyin Liang and Yubo Zhang and Yiqing Xiang and Jiaxuan Liu and Ting Sun and Manhui Lin and Yue Zhang and Changda Zhou and Tingquan Gao and Cheng Cui and Yi Liu and Dianhai Yu and Yanjun Ma},
      year={2026},
      eprint={2606.03264},
      archivePrefix={arXiv},
      primaryClass={cs.CV},
      url={https://arxiv.org/abs/2606.03264}, 
}

@article{wang2026mineru2,
  title={MinerU2. 5-Pro: Pushing the Limits of Data-Centric Document Parsing at Scale},
  author={Wang, Bin and He, Tianyao and Ouyang, Linke and Wu, Fan and Zhao, Zhiyuan and Chu, Tao and Qu, Yuan and Jin, Zhenjiang and Zeng, Weijun and Miao, Ziyang and others},
  journal={arXiv preprint arXiv:2604.04771},
  year={2026}
}

@misc{kimiteam2026kimik25visualagentic,
      title={Kimi K2.5: Visual Agentic Intelligence}, 
      author={Kimi Team and Tongtong Bai and Yifan Bai and Yiping Bao and S. H. Cai and Yuan Cao and Y. Charles and H. S. Che and Cheng Chen and Guanduo Chen and Huarong Chen and Jia Chen and Jiahao Chen and Jianlong Chen and Jun Chen and Kefan Chen and Liang Chen and Ruijue Chen and Xinhao Chen and Yanru Chen and Yanxu Chen and Yicun Chen and Yimin Chen and Yingjiang Chen and Yuankun Chen and Yujie Chen and Yutian Chen and Zhirong Chen and Ziwei Chen and Dazhi Cheng and Minghan Chu and Jialei Cui and Jiaqi Deng and Muxi Diao and Hao Ding and Mengfan Dong and Mengnan Dong and Yuxin Dong and Yuhao Dong and Angang Du and Chenzhuang Du and Dikang Du and Lingxiao Du and Yulun Du and Yu Fan and Shengjun Fang and Qiulin Feng and Yichen Feng and Garimugai Fu and Kelin Fu and Hongcheng Gao and Tong Gao and Yuyao Ge and Shangyi Geng and Chengyang Gong and Xiaochen Gong and Zhuoma Gongque and Qizheng Gu and Xinran Gu and Yicheng Gu and Longyu Guan and Yuanying Guo and Xiaoru Hao and Weiran He and Wenyang He and Yunjia He and Chao Hong and Hao Hu and Jiaxi Hu and Yangyang Hu and Zhenxing Hu and Ke Huang and Ruiyuan Huang and Weixiao Huang and Zhiqi Huang and Tao Jiang and Zhejun Jiang and Xinyi Jin and Yu Jing and Guokun Lai and Aidi Li and C. Li and Cheng Li and Fang Li and Guanghe Li and Guanyu Li and Haitao Li and Haoyang Li and Jia Li and Jingwei Li and Junxiong Li and Lincan Li and Mo Li and Weihong Li and Wentao Li and Xinhang Li and Xinhao Li and Yang Li and Yanhao Li and Yiwei Li and Yuxiao Li and Zhaowei Li and Zheming Li and Weilong Liao and Jiawei Lin and Xiaohan Lin and Zhishan Lin and Zichao Lin and Cheng Liu and Chenyu Liu and Hongzhang Liu and Liang Liu and Shaowei Liu and Shudong Liu and Shuran Liu and Tianwei Liu and Tianyu Liu and Weizhou Liu and Xiangyan Liu and Yangyang Liu and Yanming Liu and Yibo Liu and Yuanxin Liu and Yue Liu and Zhengying Liu and Zhongnuo Liu and Enzhe Lu and Haoyu Lu and Zhiyuan Lu and Junyu Luo and Tongxu Luo and Yashuo Luo and Long Ma and Yingwei Ma and Shaoguang Mao and Yuan Mei and Xin Men and Fanqing Meng and Zhiyong Meng and Yibo Miao and Minqing Ni and Kun Ouyang and Siyuan Pan and Bo Pang and Yuchao Qian and Ruoyu Qin and Zeyu Qin and Jiezhong Qiu and Bowen Qu and Zeyu Shang and Youbo Shao and Tianxiao Shen and Zhennan Shen and Juanfeng Shi and Lidong Shi and Shengyuan Shi and Feifan Song and Pengwei Song and Tianhui Song and Xiaoxi Song and Hongjin Su and Jianlin Su and Zhaochen Su and Lin Sui and Jinsong Sun and Junyao Sun and Tongyu Sun and Flood Sung and Yunpeng Tai and Chuning Tang and Heyi Tang and Xiaojuan Tang and Zhengyang Tang and Jiawen Tao and Shiyuan Teng and Chaoran Tian and Pengfei Tian and Ao Wang and Bowen Wang and Chensi Wang and Chuang Wang and Congcong Wang and Dingkun Wang and Dinglu Wang and Dongliang Wang and Feng Wang and Hailong Wang and Haiming Wang and Hengzhi Wang and Huaqing Wang and Hui Wang and Jiahao Wang and Jinhong Wang and Jiuzheng Wang and Kaixin Wang and Linian Wang and Qibin Wang and Shengjie Wang and Shuyi Wang and Si Wang and Wei Wang and Xiaochen Wang and Xinyuan Wang and Yao Wang and Yejie Wang and Yipu Wang and Yiqin Wang and Yucheng Wang and Yuzhi Wang and Zhaoji Wang and Zhaowei Wang and Zhengtao Wang and Zhexu Wang and Zihan Wang and Zizhe Wang and Chu Wei and Ming Wei and Chuan Wen and Zichen Wen and Chengjie Wu and Haoning Wu and Junyan Wu and Rucong Wu and Wenhao Wu and Yuefeng Wu and Yuhao Wu and Yuxin Wu and Zijian Wu and Chenjun Xiao and Jin Xie and Xiaotong Xie and Yuchong Xie and Yifei Xin and Bowei Xing and Boyu Xu and Jianfan Xu and Jing Xu and Jinjing Xu and L. H. Xu and Lin Xu and Suting Xu and Weixin Xu and Xinbo Xu and Xinran Xu and Yangchuan Xu and Yichang Xu and Yuemeng Xu and Zelai Xu and Ziyao Xu and Junjie Yan and Yuzi Yan and Guangyao Yang and Hao Yang and Junwei Yang and Kai Yang and Ningyuan Yang and Ruihan Yang and Xiaofei Yang and Xinlong Yang and Ying Yang and Yi Yang and Yi Yang and Zhen Yang and Zhilin Yang and Zonghan Yang and Haotian Yao and Dan Ye and Wenjie Ye and Zhuorui Ye and Bohong Yin and Chengzhen Yu and Longhui Yu and Tao Yu and Tianxiang Yu and Enming Yuan and Mengjie Yuan and Xiaokun Yuan and Yang Yue and Weihao Zeng and Dunyuan Zha and Haobing Zhan and Dehao Zhang and Hao Zhang and Jin Zhang and Puqi Zhang and Qiao Zhang and Rui Zhang and Xiaobin Zhang and Y. Zhang and Yadong Zhang and Yangkun Zhang and Yichi Zhang and Yizhi Zhang and Yongting Zhang and Yu Zhang and Yushun Zhang and Yutao Zhang and Yutong Zhang and Zheng Zhang and Chenguang Zhao and Feifan Zhao and Jinxiang Zhao and Shuai Zhao and Xiangyu Zhao and Yikai Zhao and Zijia Zhao and Huabin Zheng and Ruihan Zheng and Shaojie Zheng and Tengyang Zheng and Junfeng Zhong and Longguang Zhong and Weiming Zhong and M. Zhou and Runjie Zhou and Xinyu Zhou and Zaida Zhou and Jinguo Zhu and Liya Zhu and Xinhao Zhu and Yuxuan Zhu and Zhen Zhu and Jingze Zhuang and Weiyu Zhuang and Ying Zou and Xinxing Zu},
      year={2026},
      eprint={2602.02276},
      archivePrefix={arXiv},
      primaryClass={cs.CL},
      url={https://arxiv.org/abs/2602.02276}, 
}

\setcounter{figure}{0}
\makeatletter 
\renewcommand{\thefigure}{A\@arabic\c@figure}
\makeatother

\setcounter{table}{0}
\makeatletter 
\renewcommand{\thetable}{A\@arabic\c@table}
\makeatother

\clearpage 
\newpage

\end{document}